\documentclass{article}

\usepackage{relsize}
\usepackage{amssymb}
\usepackage{dsfont}
\usepackage{todonotes}
\usepackage{upgreek}
\usepackage[version=3]{mhchem} 
\usepackage{units}
\usepackage{bbm}
\usepackage[english]{babel}
\usepackage{titling}
\usepackage{blindtext}

\newcommand{\R}{\mathbb{R}}
\newcommand{\N}{\mathbb{N}}

\newcommand{\Prob}{\mathbb{P}}

\newcommand{\batch}{\nu}
\newcommand{\convLayer}{convolution + batch norm. + ReLu}
\newcommand{\maxPoolLayer}{max pooling}

\newcommand{\fRealized}{D_\mathrm{f}}
\newcommand{\fRealizedpredicted}{\widehat{D}_\mathrm{f}}
\newcommand{\fMean}{\overline{D}_\mathrm{f}}

\newcommand{\fModel}{\theta_{\mathrm{D_f}}}
\newcommand{\fModelpredicted}{\widehat{\theta}_{\mathrm{D_f}}}

\newcommand{\rRealized}{\rho}
\newcommand{\rRealizedpredicted}{\widehat{\rho}}
\newcommand{\rModel}{\theta_{\mathrm{\uprho}}}
\newcommand{\rModelpredicted}{\widehat{\theta}_{\mathrm{\uprho}}}

\newcommand{\cAModel}{\theta_{\mathrm{0}}}
\newcommand{\cAModelpredicted}{\widehat{\theta}_{\mathrm{0}}}

\newcommand{\cBModel}{\theta_{\mathrm{1}}}
\newcommand{\cBModelpredicted}{\widehat{\theta}_{\mathrm{1}}}

\newcommand{\finalClusterSizeB}{S_\mathrm{\ce{TiO2}}}
\newcommand{\coordinationTotal}{Z_\mathrm{total}}
\newcommand{\coordinationHetero}{Z_\mathrm{hetero}}

\newcommand{\CNNDf}{\mathrm{CNN^{\mathrm{D_f}}}}

\newcommand{\CNNreg}{\mathrm{CNN_{reg}^{\rho}}}
\newcommand{\CNNclass}{\mathrm{CNN_{class}^{\rho}}}

\newcommand{\CNNcA}{\mathrm{CNN^{0}}}
\newcommand{\CNNcB}{\mathrm{CNN^{1}}}

 \renewcommand{\arraystretch}{0.75}

\newcommand{\NA}{N_\mathrm{A}}
\newcommand{\NC}{N_\mathrm{C}}

\newcommand{\rgA}{R_\mathrm{A}}
\newcommand{\rgC}{R_\mathrm{C}}

\newcommand{\numC}{{N'}}

\author{Lukas Fuchs$^1$,
    Tom Kirstein$^1$,
    Christoph Mahr$^2$,
    Orkun Furat$^1$, \\
    Valentin Baric$^{3,4}$,
    Andreas Rosenauer$^2$,
    Lutz Mädler$^{3,4}$,
    Volker Schmidt$^1$}

\date{ \footnotesize
$^1$Institute of Stochastics, Ulm University, 
89069 Ulm, Germany\\
$^2$Institute of Solid State Physics, University of Bremen, 
28359 Bremen, Germany\\
$^3$ University of Bremen, Faculty of Production Engineering, 
28359 Bremen, Germany\\
$^4$Leibniz Institute for Materials Engineering IWT, 
28359 Bremen, Germany
}

\title{Using convolutional neural networks for stereological characterization of 3D hetero-aggregates based on synthetic STEM data}
\begin{document}
\begin{titlingpage}
    \maketitle
\begin{abstract}
    The structural characterization of hetero-aggregates in 3D is of great interest, e.g., for deriving process-structure or structure-property relationships. However, since 3D imaging techniques are often difficult to perform as well as time and cost intensive, a characterization of hetero-aggregates based on 2D image data is desirable, but often non-trivial. To overcome the issues of characterizing  3D structures from 2D measurements, a method is presented that relies on machine learning combined with methods of spatial stochastic modeling, where the latter are utilized for the generation of synthetic training data. This kind of training data has the advantage that time-consuming experiments for the synthesis of differently structured materials followed by their 3D imaging can be avoided. More precisely, a parametric stochastic 3D model 
    is presented, from which a wide spectrum of virtual  hetero-aggregates can be generated. Additionally, the virtual  structures are passed to a physics-based simulation tool in order to generate virtual scanning transmission electron microscopy (STEM) images.
    The preset parameters of the 3D model together with the  simulated STEM images  serve as a database for  the training of convolutional neural networks, which can be used  to determine the parameters of the underlying 3D model and, consequently, to predict 3D structures of hetero-aggregates from 2D STEM images. 
    Furthermore, an error analysis 
    is performed to evaluate the prediction power of the trained neural networks with respect to structural descriptors, e.g. the hetero-coordination number.   

\vspace{1cm}
    
    \noindent\textbf{Keywords:} synthetic HAADF-STEM, nanoparticle aggregate, hetero-aggregate, convolutional neural network, stereological characterization, stochastic 3D model, statistical image analysis
    
\end{abstract}

\end{titlingpage}

\section{Introduction}\label{section:introduction}
The properties of many functional materials depend to a large extent on their structure and chemical composition. Hence, measuring both is mandatory in order to understand and improve their effective properties. An important class of materials are hetero-aggregates, which are compositions of at least two dissimilar classes of primary particles, called, for the sake of simplicity,  particles from now on. Properties of hetero-aggregates can be quite different in comparison to aggregates that consist of monodisperse particles. A prominent example in applications concerned with photocatalysis are hetero-aggregates made of titanium dioxide (\ce{TiO2}) and tungsten trioxide (\ce{WO3})~\cite{PinedoEscobar2021,Kwon2000, Yan2012}. The combination of both materials leads to aggregates with hetero-junctions, i.e., points at which two particles made from different materials touch. At such junctions, photogenerated electron-hole pairs are spatially separated, 
hindering their direct recombination, which results in a higher photocatalytic activity compared to pure \ce{TiO2}~\cite{Low2017}.

In order to accurately investigate the properties of hetero-aggregates with imaging techniques, it is essential to resolve the individual  particles within the structure. A suitable tool for the characterization of hetero-aggregates, consisting of particles with radii of a few nanometers, is (scanning) transmission electron microscopy, (S)TEM. With a spatial resolution in the sub-nanometer regime, even the atomic structure can be investigated. However, in conventional STEM only two-dimensional (2D) projection images of the aggregates can be acquired while information about the third dimension is lost. This problem can be overcome using STEM tomography, where the sample is tilted with respect to the electron beam such that a series of projection images under various projection angles is acquired, see~\cite{Midgley2003}. From this series of STEM projection images, the three-dimensional structure can be reconstructed, e.g., with iterative reconstruction techniques~\cite{Gilbert1972}. The major disadvantage of STEM tomography is the fact that acquisition of a single tilt series can take several hours, and thus, this method does hardly allow for the investigation of a large number of aggregates. Furthermore, many samples do not allow for such a long measurement as hetero-aggregates and nanoparticles can change their structure and arrangement during extensive exposure to the electron beam, hindering the reconstruction.

As opposed to STEM tomography, 2D STEM images can be acquired within a few seconds, allowing for the acquisition of several images of various aggregates in a reasonable amount of time. For this reason, it is desirable to use 2D STEM images in order to characterize the  3D morphology  of aggregates. This can be
achieved by training  neural networks to predict structural properties of 3D hetero-aggeregates from 2D STEM images. However, the training of  neural networks requires a broad database of pairs of differently structured hetero-aggregate and corresponding  2D STEM images. The experimental acquisition of such a database, i.e., the synthesis of differently structured aggregates and their imaging
would be expensive in both time and resources. Alternatively, simulated image data can be used for training purposes, see \cite{frei2020image} for a similar approach.

In the present paper, a stochastic 3D model for the generation of virtual aggregates and a physics-based STEM model for the simulation of corresponding 2D  STEM images is combined in order to provide training data. In other words, methods of stochastic geometry \cite{chiu.2013} are utilized to derive a parametric model for the generation of a wide spectrum of virtual, but realistic  aggregates.
 Additionally, the virtual structures are passed to a physics-based simulation tool in
order to generate virtual scanning transmission electron microscopy (STEM)
images. The preset parameters of the 3D model together with the simulated STEM images serve as a database for the training of convolutional neural
networks, which can be used to predict the parameters of the underlying 3D
model and, consequently, to predict 3D structures of hetero-aggregates from
2D STEM images.
In literature, there are already CNN-based approaches that do not use stochastic geometry models to generate stochastically equivalent 3D structures from 2D images~\cite{kench2021generating}. However, the presented approach aims to generate such digital shadows by combining a well-established parametric stochastic 3D model and a CNN-based approach. In order to use such a parametric stochastic 3D model to generate digital shadows of hetero-aggregates, appropriate values of the model parameters must be chosen. The focus of the present paper is on this calibration procedure, also called model fitting.

More specifically, it is investigated how convolutional neural networks (CNNs) \cite{Goodfellow2016,James2021} can be used to determine the parameters of the stochastic 3D model and, consequently,  to generate digital shadows of 3D aggregates, from  2D STEM images.
CNNs are a type of artificial neural networks commonly used in image analysis and recognition tasks, see e.g.~\cite{ma16093397}.
They consist of multiple layers of neurons that learn to recognize patterns and features in the input data through a calibration process, called training.

In conventional spatial stochastic modeling of complex 3D morphologies, the process of model fitting  typically involves several steps, see for example~\cite{NEUMANN2023112394,weber2023modeling}. First, image data has to be acquired, preprocessed, and segmented. Subsequently, an appropriate model type is chosen, and its model  parameters are adjusted accordingly using descriptive statistics of the segmented image data. However, the approach considered in the present paper differs from the classical one. On the one hand, the image data does not have to be segmented, which is advantageous since image segmentation can be a time-consuming complex task. Moreover, the model parameters are predicted by the neural networks directly, meaning that the  descriptive statistics 
are not chosen by hand. This allows for the use of stochastic 3D models with parameters which are not easily estimatable from the image data.

In order to evaluate the performance of such a CNN-based approach, structural descriptors of aggregates drawn from the stochastic 3D model with preset parameter values are compared with structural descriptors of aggregates  drawn from the 3D model with parameter values predicted by the CNN-based approach.

However, the structural similarity of the measured image data of  aggregates and image data drawn from the fitted 3D model  strongly depends on two factors, (i) the suitability of the  chosen model type  for the given data, and (ii) the ability of the selected CNN approach to determine the parameters of the stochastic 3D model from 2D STEM image data.
More specifically, when analyzing measured image data of experimentally synthesized hetero-aggregates,  there might not be any configuration of model parameters  that results in a high-quality fit.
In this case, the dissimilarities between the original image data and its digital shadows, generated by the fitted model, cannot necessarily be attributed to the fitting procedure, but rather to the inadequate  choice of the model type. 
Thus, in the present paper, to be able to attribute these dissimilarities to an inadequate CNN approach, including data preprocessing, model architecture and learning procedure, 
the same stochastic  3D model is used as both the generator for the training data and the model to be fitted. 



For an adequately designed CNN approach and adequately chosen type of the stochastic 3D model,
 the digital shadows drawn from the fitted 3D model should be statistically equivalent to  experimentally synthesized aggregates in terms of their 3D structure and chemical composition. Then,
these digital shadows can be used as geometry input of (spatially resolved) numerical modeling and simulation,  to determine their functional properties,  see e.g. \cite{neumann2016stochastic,prifling2021generating}. In this way, 3D imaging techniques like STEM tomography of the aggregates can be avoided  in order to derive quantitative process-structure or structure-property relationships for hetero-aggregates. 
Note that by means of such relationships, optimized specifications of process parameters  can be deduced, which lead to hetero-aggregates with desired structures and properties. 
Digital shadows used for  structure-property optimization are also referred to as digital twins. Their implementation  will be the subject of a forthcoming study.

The present work is organized as follows: In Section~\ref{sec.two} the CNN-based approach is described to   predict the 3D structure of hetero-aggregates from 2D STEM images. In particular, the generation of synthetic training data is explained which are used for the prediction of model parameters. Then, in Section~\ref{sec.three}, the results are presented which have been obtained for various aspects of model parameter prediction. Section~\ref{sec.four} compares the methods developed in the present paper  with analysis tools considered in the literature. Section~\ref{sec.five} concludes.

\section{Methods}\label{sec.two}

\begin{figure}[ht]
    \centering
    \includegraphics[width=1\textwidth]{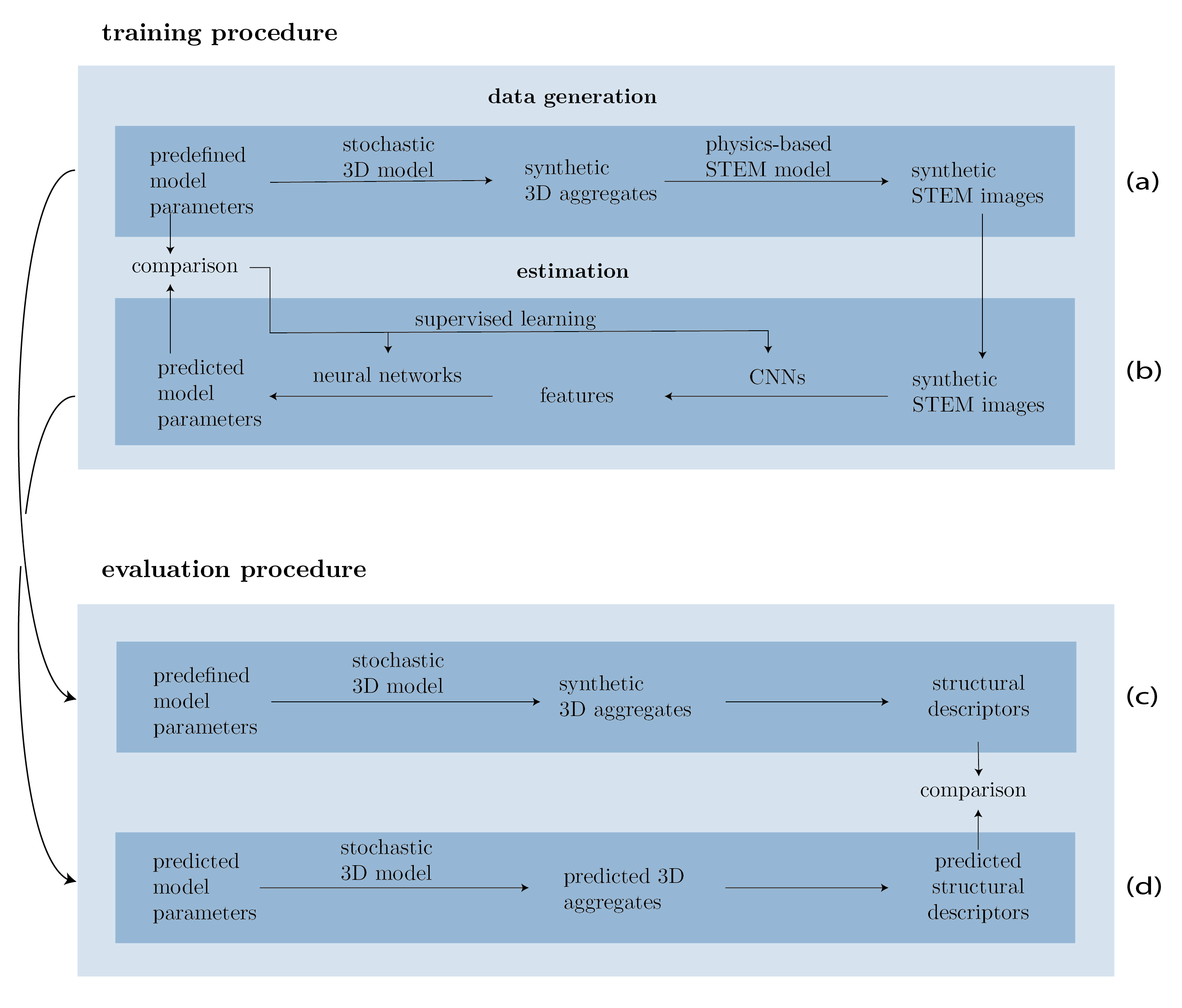}
    \caption{Workflow  of the training and evaluation procedure.}
    \label{figure:sketch_procedure}
\end{figure}

This section provides details how the presented CNN-based approach is built for predicting the 3D structure  of hetero-aggregates from 2D STEM images.
It comprises two main steps. First, virtual but realistic STEM images are generated from simulated 3D image data. 
More specifically, synthetic  aggregates are drawn from a stochastic 3D model  with preset model parameters, where the latter describe the aggregation procedure simulated by the model and therefore influence structural properties of the generated aggregates.
These aggregates are then used to generate corresponding STEM images by means of  a physics-based simulation tool, see Figure~\ref{figure:sketch_procedure}a. Systematically varying the parameters of the stochastic 3D model provides a wide range of differently structured  aggregates and their  STEM images. In a second step, visualized in Figure~\ref{figure:sketch_procedure}b,  the parameters of the stochastic 3D model together with the  simulated STEM images  serve as a database for  the training of   CNNs, in order to learn how to reconstruct the  parameters of the stochastic 3D model  from  STEM images. 
For the reconstruction, initially, a CNN extracts features from STEM images which characterize the depicted structure of aggregates in an informative but not necessarily interpretable manner. Then, these features are utilized to predict our interpretable predefined model parameters. For more details, see Section~\ref{sec:cnn}. This approach is designed to allow for quick and accurate prediction of model parameters for real  hetero-aggregates from measured STEM images  and, consequently, to predict the 3D
morphology of hetero-aggregates from 2D STEM images.

The quality of the predictor is evaluated with respect to the similarity between predefined and predicted  model parameters. Recall that interpretable model parameters describe the aggregation procedure simulated by the stochastic model. Thus, a good match between predefined and predicted model parameters can already be an indication for a good structural match between aggregates generated by the model with predefined/predicted parameters. Nevertheless, some structural descriptors (i.e., quantities which characterize the structure of aggregates like hetero-coordination number) may be sensitive with respect to changes in the model parameters.

Therefore, the quality of the predictor is further evaluated 
by comparing structural descriptors of  aggregates drawn from  stochastic 3D models with  pre-defined and predicted parameters, respectively, see Figure~\ref{figure:sketch_procedure}c,d. The structural descriptors considered in this paper, which are chosen due to their relevance in  process engineering, are displayed in Table~\ref{tab::list_descriptors}.
They are complementary to the features, utilized in the model parameter prediction. Furthermore, these descriptors are interpretable and characterize the 3D structure of the aggregates (whereas the features describe the structure observed in 2D images).


 \begin{table}[ht]
        \footnotesize
        \centering
        \begin{tabular}{|l|l|} 
        \hline
         descriptor & symbol\\
         \hline
         average cluster sizes of \ce{TiO2} particles&$\finalClusterSizeB$\\
         hetero-coordination number &$\coordinationHetero$ \\
         coordination number &$\coordinationTotal$ \\
         \hline
        \end{tabular}
        \caption{Structural descriptors used for evaluating the model parameter prediction. For formal definitions of the descriptors, see Section~\ref{sec::structural_descriptors}.}
        \label{tab::list_descriptors}
    \end{table}

\subsection{Generation of synthetic training data}\label{sec.two.one}
The use of synthetic training data requires careful attention to ensure that the artificially generated data accurately reflects particularities of experimentally measured data such that a regression model (e.g., a CNN) trained on synthetic data can be extended to new, real-world data. More precisely, if the generation of realistic data is successful, a network trained on this data can be used for applications on real-world data, and thus, reducing the amount of experimentally measured and labeled training data.


In the present study, synthetic training data was generated through a three-step process. First, virtual hetero-aggregates were generated using a stochastic 3D model. Then, using a physics-based simulation tool, STEM intensities were determined based on the material and thickness of the aggregates. Finally,  virtual but realistic STEM images were computed by adding noise and other sources of variability to the previously determined STEM intensities. In the following, the stochastic 3D model is introduced and then more details about each of the data generation steps mentioned above is provided.

\subsubsection{Stochastic 3D model}\label{sub.sto.mod}
In this section, the stochastic 3D model is introduced, which will be used to generate a wide spectrum of virtual hetero-aggregates by varying the values of four different model parameters, denoted by $\fModel, \rModel, \cAModel,\cBModel$, where $\fModel\in(1,3), \rModel\in(0,1)$, and $\cAModel,\cBModel   \in\N=\{1,2,\ldots\}$.

These model parameters  control the fractal dimension, the mixing ratio, and  clustering properties of the hetero-aggregates, respectively.

Throughout this paper, a spherical particle is defined as a triplet $p=(x,r,l)$ of particle position $x \in \R^3$, radius $r \in\R^+=(0,\infty)$ and label $l\in \{0,1\}$. Moreover, a hetero-aggregate $A$, consisting of $N$ particles for some fixed  $N \in \N$, is a set of connected and non-overlapping spherical particles, i.e., 
\begin{align}
    A= \{p_i=(x_i,r_i,l_i):x_i \in \R^3, ~ r_i \in\R^+,~ l_i \in  \{0,1\},~ 1\leq i\leq N \}.\label{def:rep_aggregat}
\end{align}
In this context, two particles $p,p^\prime \in A$ are said to be connected if for some $j\in\{2,\ldots,N\}$  there is a set of indices $\{i_1,\ldots,i_j\} \subset \{1, \ldots, N\}$ with $p=p_{i_1}$ and $p^\prime=p_{i_j}$, such that 
\begin{align}
    \|x_{i_k}-x_{i_{k+1}}\|\leq 1.01(r_{i_k}+r_{i_{k+1}}) ~\text{ for all } k\in\{1,\ldots,j\}\,,\label{def.con.con}
\end{align}
where $\| y \| = \sqrt{\sum_{k=1}^{3} y_k^2} $ denotes the   Euclidean norm of \mbox{$y=(y_1, y_2, y_3) \in\R^{3}$}. The prefactor $1.01$ in Eq.~(\ref{def.con.con}) represents the maximum distance of particles which is allowed to  consider them to be in contact. It is determined to be $1\%$ of the sum of their radii.
Moreover, two particles $p=(x,r,l),p^\prime=(x^\prime,r^\prime,l^\prime)$ are said to be  overlapping if the distance of their centers is smaller than the sum of their radii, i.e., $\|x-x^\prime\|< r+r^\prime$.
The label $l$ of a particle $p=(x,r,l)$ determines its material. More precisely, in our case, a particle with label $l=0$ consists of $\ce{WO3}$, whereas a particle with label $l=1$ consists of $\ce{TiO2}$.

The mixing ratio $\rho$ of an aggregate $A$ is defined as its fraction of particles with label $l=0$, i.e.,
\begin{align}
\rho(A)=\frac{\#\{p_i\in A \colon l_i=0\}}{\#A}\,,
\label{def.rho}
\end{align}
where $\#$ denotes cardinality.

Notice the distinction in notation between $\rModel$ and $\rho$ since these values are not necessarily equal. More precisely, the model parameter $\rModel$ can be set to an arbitrary value in the interval $[0,1]$ and it primarily influences the distribution of the structural descriptor $\rho$ of aggregates generated with $\fModel$, as explained in more detail later on.
Furthermore, the radius of gyration $R_\mathrm{g}>0$ of 
an aggregate $A$ is given by
\begin{align}\label{eq:r_gyr}
R_\mathrm{g} &= \sqrt{\frac{\sum_{i=1}^{N}m_i\cdot \|x_i - c_0\|^2}{\sum_{i=1}^N m_i}} ~, \quad\text{with} ~~ c_0 = \frac{ \sum_{i=1}^N m_i x_i}{\sum_{i=1}^N m_i}\,,
\end{align}
where $m_1, \dots, m_N>0$ denote the particle masses and $c_0$ is the aggregate's center of mass. 

The stochastic 3D model described below is motivated by the idea that hetero-aggregates have a fractal-like structure~\cite{Forrest1979, Meakin1983}. This fractal-like structure of an aggregate $A$ can be quantified by the so-called fractal dimension $D_\mathrm{f}$, given by
\begin{align}
D_{\mathrm{f}} = \frac{\log\left(\frac{N}{k_{\mathrm{f}}}\right)}{\log\left(\frac{R_{\mathrm{g}}}{a}\right)}\,,
~\label{Eq::FractalDimension}
\end{align}  
\noindent
where $k_\mathrm{f}>0$ is a fractal prefactor, which  is set  to $1.3$, and $a=\frac{1}{N}\sum_{i=1}^N r_i$ is the mean radius of the particles.
For example, aggregates with a fractal dimension $D_\mathrm{f}$ close to $1$ are arranged in a nearly straight line, whereas those with a fractal dimension $D_\mathrm{f}$ close to $3$ are composed of densely packed particles. Thus, realistic hetero-aggregates have values for $D_\mathrm{f}$ within the interval $(1,3)$, see e.g.~\cite{Cai1995, Eggersdorfer2012, Eggersdorfer2014,Filippov2000, Forrest1979}.

Note that the  hetero-aggregate model presented in this paper is based on cluster-cluster aggregation, which involves a two-stage process for aggregate formation. In the first stage, primary particles aggregate to form small, homogeneous primary clusters. These primary clusters then undergo a second aggregation stage, leading to larger hetero-aggregates.

If an aggregate is homogeneous, i.e., all its particles share the same material, the labels $\{l_i\}_{i=1}^N$ will be neglected, and therefore the description of the aggregate $A$ can be compressed to
\begin{align}
    A= \{p_i=(x_i,r_i):x_i \in \R^3, ~ r_i \in\R^+,~ 1\leq i\leq N \}.\label{def:rep_homo_aggregat}
\end{align}
In this case, the primary cluster model  is introduced as a random set $\Phi_N= \{P_i: 1 \leq i\leq N \}\subset \R^3\times\R^+$ which models the geometry of small homogeneous clusters of size $N$ for some fixed  $N \in \N$, compare~\cite{Baric2018} for earlier work. Here,  $P_i=(X_i,R_i)$, where
 $X_i$ is a random vector and $R_i$ is  a non-negative random variable describing the position and radius of a particle, respectively, for each $i \in \{1,\ldots,N\}$.

The random variables  $R_1,\ldots,R_N$ are independent and log-normally distributed  with parameters $\mu =\unit[12]{nm}$  and $\sigma = \unit[3]{nm}$. However, the random vectors  $X_1,\ldots,X_N$  which describe the particle positions, are recursively defined due to the dependency of $X_i$ on $X_1, \ldots, X_{i-1}$ and $R_1, \ldots, R_i$ for all $1<i\leq N$. This approach ensures that every realization of $\Phi_N$ is a set of  connected and non-overlapping particles, with a predetermined fractal dimension  $D_\mathrm{f}$.
Note that  for technical reasons the random vector $X_i$ can take not only values from $\R^3$, but also the fictitious value $\infty$. The latter value is used to model invalid particle positions.

More precisely, $X_1=(0,0,0)$ and,
under the condition that the values $x_1,\ldots,x_i$ and $r_1,\ldots,r_{i+1}$ of $X_1,\ldots,X_i$ and $R_1,\ldots,R_{i+1}$ are given for some $i\in\{1,\ldots,N-1\}$,   the random vector $X_{i+1}$
 is uniformly distributed on some set $L(A,r_{i+1})\subset\R^3$,  provided   that  $(\infty,r) \not \in A $ for all $ r \in \R^+$ and $L(A,r_{i+1}) \not = \emptyset$, otherwise $X_{i+1}=\infty$. 
 Here,  $A=\{(x_1,r_1),\ldots,(x_i,r_i)\}$ and $L(A,r_{i+1}) \subset \R^3$ is  the set of all permissible particle positions $x\in\R^3$ such that the set $A \cup\{(x,r_{i+1})\}$ describes a cluster of connected and non-overlapping particles with fractal dimension $D_\mathrm{f}$
 being equal to some  preset value $\fModel\in(1,3)$.
  In other words, $L(A,r_{i+1})$ is the set of positions where a particle of radius $r_{i+1}$ can be added to the cluster $A$ without violating
  the equation  $D_\mathrm{f}=\fModel$.  If no such position exists, $X_{i+1}$ will be assigned $\infty$, indicating that the cluster $A$ cannot 
  be extended.

To draw a sample from the random set $\Phi_N= \{P_i: 1 \leq i\leq N \}\subset \R^3\times\R^+$, the procedure described above
is repeated until $X_i\not=\infty$ for all $i=1,\ldots,N$.  
The primary clusters generated in this way then undergo a second aggregation stage, leading to larger, hetero-aggregates which consist of $\numC$ primary clusters for some integer $\numC\in\N$.

More formally, for some sequence of primary cluster sizes $N_1,\ldots,N_\numC$, $\numC$ independent random sets  $\Phi^{(1)}_{N_1},\ldots, \Phi^{(n)}_{N_\numC}$ are considered as described above.
 The cluster $\Phi^{(k)}_{N_k}$
  is assigned a random position $C_k$ in $\R^3\cup\{\infty\}$ for each $k\in\{1,\ldots,\numC\}$, ensuring that realizations of the resulting hetero-aggregates are union sets of connected and non-overlapping spheres, which adhere to a preset fractal dimension $D_\mathrm{f}=\fModel$. In the following,
the cluster $\Phi_{N_k}^{(k)}$ which has been shifted by a (random) displacement vector $C_k$ is denoted  by  $\Phi_{N_k}^{(k)} + C_k = \{(X+C_k,R) \colon (X,R) \in \Phi_{N_k}^{(k)} \}$. 
Furthermore, for each $k\in\{1,\ldots,\numC\}$,
the cluster
$\Phi_{N_k}^{(k)} + C_k$ is assigned a (random) label $L_k$ which can be equal to $0$ or $1$, determining whether the cluster consists of \ce{WO3}  or  \ce{TiO2}.
The clusters of label 0 have a size of $\cAModel$ whereas the clusters of label 1 have a size of $\cBModel$.
These cluster sizes $\cAModel,\cBModel \in \{1,\ldots,6\}$ are a further model parameter. The labeled version of $\Phi_{N_k}^{(k)}+C_k$ with the (random) label $L_k$ is denoted by $(\Phi_{N_k}^{(k)}+C_k) \times L_k = \{(X+C_k,R,L_k) \colon (X,R) \in \Phi_{N_k}^{(k)} \}$. 
Finally, the stochastic
3D model  $\Psi_\numC$ 
of hetero-aggregates, which consist of $\numC$ primary clusters, is  given by 
\begin{align}
    \Psi_\numC = \bigcup_{k=1}^\numC(\Phi^{(k)}_{N_k} + C_k) \times L_k.\label{def.psi.enn}
\end{align}
 Here, the random variables $L_1,\ldots,L_\numC$, modeling the labels of the primary clusters, are independent and Bernoulli-distributed with $\Prob(L_k=1)=\frac{(1-\rModel) \cAModel}{(1-\rModel) \cAModel+\rModel \cBModel}$ for each $k\in\{1,\ldots,\numC\}$. 
 Note that the label of a primary cluster does not only determine its material but also its size. Specifically, the  size $N_k$ of the  $k$-th primary cluster  is given by $N_k = \cAModel +L_k (\cBModel-\cAModel)$  for each $k\in\{1,\ldots,\numC\}$, i.e., a cluster has a size of $\cAModel$ if its label is equal to $0$, and $\cBModel$ otherwise. For sufficiently large $\numC\in\N$, according to the law of large numbers, these definitions of $L_1,\ldots,L_\numC$ and  $N_1,\ldots,N_\numC$ ensure that the mixing ratios $\rho$ of  hetero-aggregates drawn from the stochastic 3D model
 $\Psi_\numC$ are approximately equal to the preset value $\rModel$.

 The random displacement vectors $C_1,\ldots,C_\numC$ that describe the  positions of primary clusters in the hetero-aggregate model $\Psi_\numC$ are again defined recursively to ensure that the particles of the random hetero-aggregate are connected and non-overlapping and that the fractal dimension $\fModel$ is maintained. More precisely, $C_1$ is put to $(0,0,0)$ and, given that $\bigcup_{k=1}^i(\Phi^{(k)}_{N_k} + C_k) =A_1$ and
$\Phi_{N_i}^{(i+1)}=A_2$ 
for some $i\in\{1,\ldots,\numC-1\}$,
the random vector $C_{i+1}$ is  
  uniformly distributed on some set $\widetilde{L}(A_1,A_2)\subset\R^3$, provided that  
  $(\infty,r) \not \in A_1 \cup A_2 \text{ for all } r \in \R^+$ and $\widetilde{L}(A_1,A_2) \not = \emptyset$, otherwise $C_{i+1}=\infty$. In this context, $\widetilde{L}(A_1,A_2) \subset \R^3$ is the set of all cluster positions $c\in\R^3$ for which the set $A_1 \cup (A_2  +c)$ represents a hetero-aggregate  of connected and non-overlapping particles with fractal dimension $D_\mathrm{f}$
 being equal to the  preset value $\fModel\in(1,3)$. 

The resulting hetero-aggregate model $\Psi_\numC$ which is described by the  model parameters $\fModel,\rModel,\cAModel,\cBModel$ can now be used to generate virtual aggregates. These aggregates consist of $\numC$ primary clusters with a fractal dimension of $\fModel$, an expected mixing ratio $\rModel$, and have a label-dependent clustering properties mainly influenced by $\cAModel$ and $\cBModel$.
Moreover, the model parameters have a multivariate influence on further structural descriptors, e.g. the hetero-coordination number, see Section~\ref{sec::structural_descriptors}. 
In theory, this can be achieved by drawing samples from $\Psi_\numC$ under the condition that $(\infty,r) \not \in \Psi_{\numC} $ for all $r\in R^+$. However,  due to computational limitations, this procedure can only be performed in an approximate sense. In the following section,  this  will be explained in detail.

\subsubsection{Generation of virtual hetero-aggregates}\label{section:Artificial aggregate generation}
The recursively defined models of  primary clusters and  hetero-aggregates described above can be used to construct algorithms for drawing samples from these models.
More precisely, the simulation starts by selecting an initial particle (or cluster), to which particles (or clusters) are added sequentially. Each additional particle (or cluster) is assigned a random radius (or label) and placed at a uniformly  sampled random position in $L$ (or $\widetilde{L})$, to be added to the existing cluster (or aggregate).
This procedure is iterated until a desired cluster (or aggregate) size is reached.
In the following, the desired size of each aggregate is independent, uniformly selected from the range $\{20, \ldots,80\}$.

The sets $L,\widetilde{L} \subset \R^3$ in the stochastic 3D model, from which particle (or cluster) positions are uniformly sampled in order to generate aggregates with a given fractal dimension, are only implicitly defined. Therefor, uniform sampling on $L$ and $\widetilde{L}$ is computationally expensive.

\begin{figure}[ht]
    \centering
    \includegraphics[width=1\textwidth]{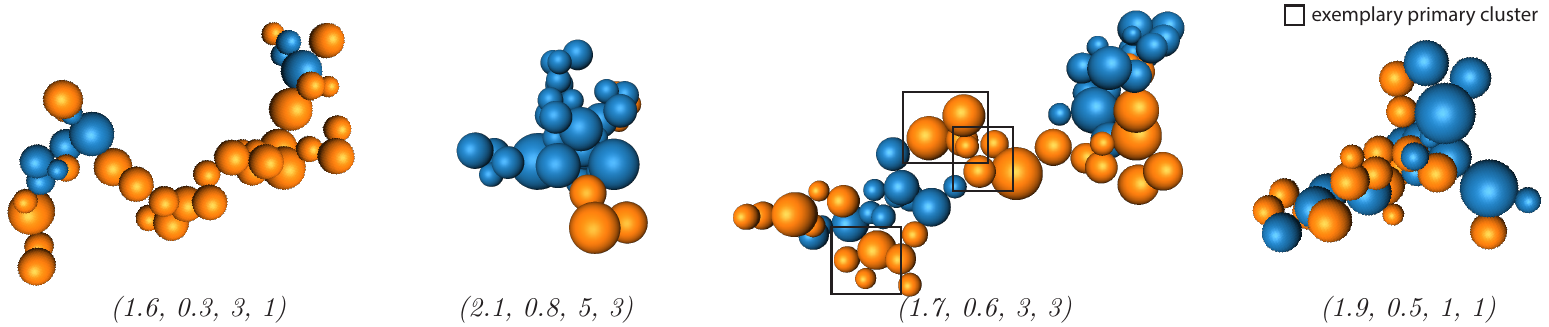}
    \caption{Examples of virtual hetero-aggregates. The labels correspond to the values of the vector  $\theta = (\fModel,\rModel,\cAModel,\cBModel)$ of their model parameters. The \ce{WO3} particles are displayed in blue, while the \ce{TiO2} particles are displayed in orange. The particle sizes are drawn from a log-normal distribution with parameters $\mu$ and $\sigma$ as defined above. Some primary clusters, as defined in Section~\ref{sub.sto.mod}, are highlighted.}
    \label{figure:different_cluster_sizes}
\end{figure}

To enable efficient uniform sampling from both $L(A, r_{i+1})$ and $\widetilde{L}(A_1, A_2)$, the radii of the particles in the sets $A \cup \{p_{i+1}\}$ and $ A_1 \cup A_2$ are temporarily replaced by their respective arithmetic mean. Note that this replacement is used exclusively when calculating the fractal dimension $\fRealized$ within the definitions of $L$ and $\widetilde{L}$.
Thus, all permissible positions for the center of mass of the added particle (or cluster) are located on the surface of a sphere around the center of mass of the cluster (or aggregate), the  radius $d$ of which is given by
\begin{align}    d=\sqrt{\frac{a^2(\NA+\NC)^2}{\NA \NC} \left (\frac{\NA+\NC}{k_f} \right )^{\frac{2}{\fRealized}}-\frac{(\NA+\NC)}{\NC}R_A^2-\frac{(\NA+\NC)}{\NA}R_C^2},
\end{align}
where $\NA$ and $\NC$ denote the number of particles in the aggregate and the cluster to be added, respectively,  $\rgA$ and $\rgC$ are their respective radii of gyration, introduced in~\ref{eq:r_gyr}, and $a$ and $k_f$ are the quantities used in the definition of $D_{\mathrm{f}}$ 
given in Eq.~\eqref{Eq::FractalDimension},
see also~\cite{Filippov2000}.
Since uniform sampling on the sphere surface can be performed efficiently, by means of rejection sampling, uniform sampling from the modified sets $L$ or $\widetilde{L}$ can be done much faster.%
This procedure results in aggregates with fractal dimensions randomly distributed around the target value $\fModel$. For further details on the distribution of the fractal dimension $\fRealized(A)$ of an aggregate $A$ generated by this model, see Section~\ref{Data labeling}. 
\noindent
For  data acquisition the four model parameters $\fModel, \rModel, \cAModel,\cBModel$ of the stochastic hetero-aggregate model are systematically varied, by name, the parameters regarding the fractal dimension $\fModel$, the intended mixing ratio $\rModel$ and the primary cluster sizes $\cAModel$ and $\cBModel$ of the two materials.
In this manner a broad spectrum of aggregates is obtained, which differ not only in preset model parameters used for their generation but also in structural descriptors like the ones listed in Table~\ref{tab::list_descriptors}. 
The fractal dimension of \ce{TiO2}-\ce{WO3} hetero-aggregates, which form by diffusion-limited cluster-cluster-aggregation, is expected to approach the value of $D_\mathrm{f}=\unit[1.5]{}$ for particles with dispersed sizes~\cite{Eggersdorfer2012},
 and $D_\mathrm{f}=\unit[1.78]{}$ for monodispersed particles~\cite{Julien1984}.
Furthermore, the fractal dimension is expected to increase, when particles start to sinter at their contact points~\cite{Eggersdorfer2014}.
In order to create a large database of differently structured virtual hetero-aggregates and their corresponding STEM images, the model parameter $\fModel$ was varied in the present work from $\fModel=1.5$ to $2.5$ in steps of $0.1$. The intended mixing ratio $\rModel$ was varied from $\rModel=0.1$ to $0.9$ in steps of $0.1$ and the primary cluster sizes $\cAModel$ and $\cBModel$ were chosen between one and six in steps of one for both materials, see also Table~\ref{label_dist}.
Some examples of virtual hetero-aggregates for various values of the model parameters $\fModel,\rModel,\cAModel,\cBModel$ are visualized in Figure~\ref{figure:different_cluster_sizes}.

 \begin{table}[ht]
        \footnotesize
        \begin{tabular}{l|c|c|c|c} 
         parameter & $\fModel$ & $\rModel$ & $\cAModel$ & $\cBModel$ \\
         \hline
         range &\{1.5, 1.6, …, 2.5\} & \{0.1, 0.2, …, 0.9\}&\{1, 2, …, 6\}&\{1, 2, …, 6\} \\
        \end{tabular}
        \caption{Range of model parameters. The model parameters $\fModel$ and $\rModel$ affect the fractal dimension and mixing ratio of the resulting aggregates, while the model parameters $\cAModel$ and $\cBModel$  determine the clustering behavior of the materials within the aggregates, respectively.}
        \label{label_dist}
    \end{table}

\subsubsection{Simulation of STEM intensities}

After generating virtual  hetero-aggregates, reference simulations to calculate the high-angle annular darkfield (HAADF)-STEM intensity of \ce{TiO2} and \ce{WO3} are conducted as a function of the sample thickness and material density.
For that purpose, multi-slice simulations in the frozen-lattice approach~\cite{VanDyck2009} with the STEMSIM software~\cite{Rosenauer2008} were performed. Simulations were done for the rutile and anatase phases of \ce{TiO2} as well as for gamma and delta phases of \ce{WO3}. Crystal parameters and Debye-Waller factors were taken from~\cite{Diehl1978, Horn1972, Loopstra1969, Sugiyama1991}, and elastic atomic scattering amplitudes from~\cite{Kirkland1998} were used. The HAADF-STEM intensity for microscope parameters equal to those one would use in experiments with a ThermoFisher 60/300 Spectra microscope were simulated. This machine is equipped with a Cs-corrector for the probe forming system, an X-FEG and SuperXG2 EDXS detectors. A semi-convergence angle of $\beta=\unit[21.1]{mrad}$ and an acceleration voltage of $\unit[300]{kV}$ were set. The simulated HAADF-STEM intensity was obtained by integration of electrons scattered into the annular range between $\unit[55]{mrad}$ and $\unit[250]{mrad}$ after application of a detector specific sensitivity curve~\cite{Rosenauer2008}.

The HAADF-STEM intensity further depends on the orientation of the crystal with respect to the electron beam. To account for this effect, various orientations for each material and phase of the crystal were simulated.

Therefore, the crystal was systematically tilted in nine equal steps from a [100]- towards a [010]-viewing direction.
In addition, a random tilt was simulated. The final result is a data set with the HAADF-STEM intensity as a function of the sample thickness for \ce{TiO2} and \ce{WO3}, each in two different crystal phases, each with ten orientations of the crystal with respect to the electron beam.




\subsubsection{Generation of realistic STEM images}
\begin{figure}[ht]
            \centering
            \includegraphics[width=\textwidth]{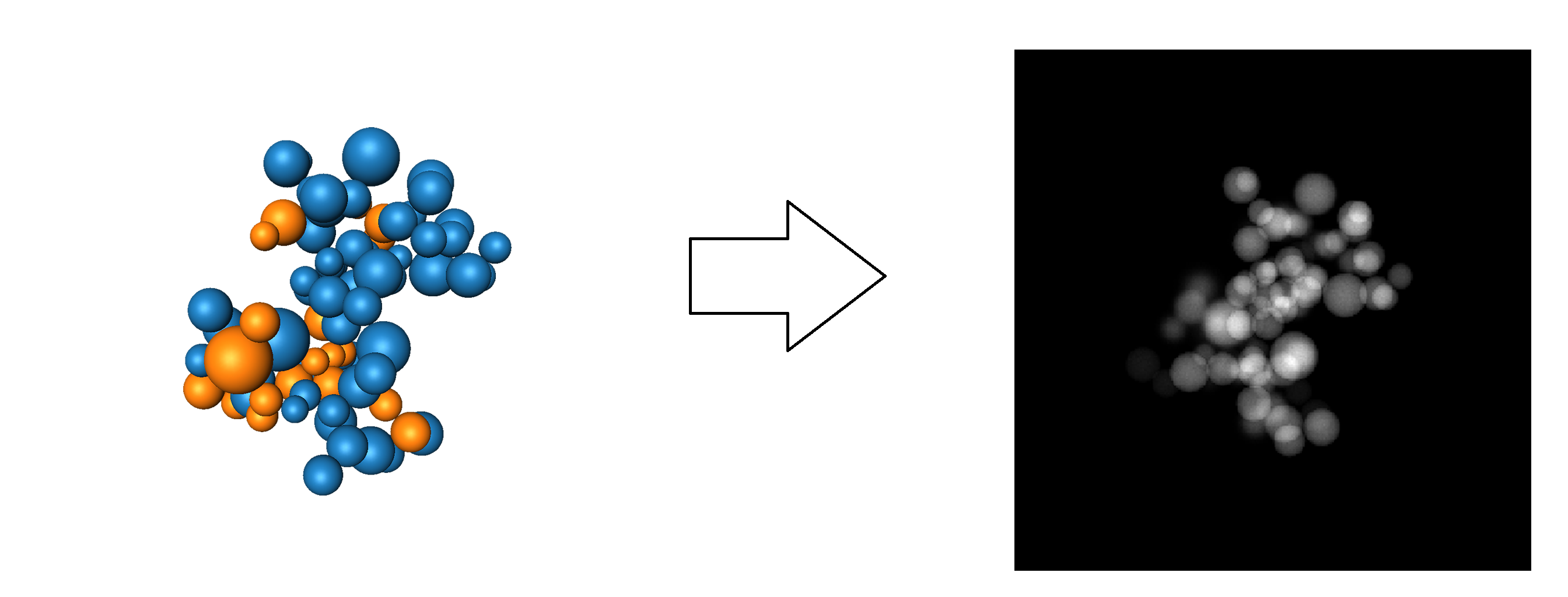}
            
            \caption{
            Schematic representation of the 3D structure of a virtual hetero-aggregate (left) and its respective STEM image (right). The \ce{WO3} particles are colored blue and correspond to the bright particles in the STEM image. The \ce{TiO2} particles are colored orange and correspond to the dark particles.}
            \label{figure_STEM_generation}
        \end{figure}

The third step combines the HAADF-STEM reference simulations described above with the virtual 3D hetero-aggregates. STEM images show 2D projections of the aggregates, see  Figure~\ref{figure_STEM_generation}.  Therefore, the projections of the individual particles along one direction are computed, as usual in electron microscopy, the electron beam direction and hence the projection direction is referred to as $z$-direction. This results in thickness maps for the individual particles. Using the reference simulations, these thickness maps are translated into maps of the HAADF-STEM intensities. To this end, for each particle, the reference simulation of the respective material was chosen in a random phase and a random orientation of the crystal with respect to the electron beam.

In an aggregate, which extends several tens of nanometers in $z$-direction, not all particles appear in focus. Only particles with centers located at height $z=\unit[0]{nm}$ are in focus as the electron beam is focused on this plane. To account for this effect, each HAADF-STEM map of the individual particles is convolved with a Gaussian kernel.
More precisely, for a particle located at height $z$, the standard deviation $\sigma_{\mathrm{STEM}}$ of the Gaussian kernel with which the corresponding HAADF-STEM map is convoluted is chosen as $\sigma_{\mathrm{STEM}} = |z|\cdot \tan(\beta)$, where $\beta=\unit[21.1]{mrad}$ is the semi-convergence angle, assuming a conical beam shape.
Then, blurred HAADF-STEM maps of individual particles are summed up to obtain the artificial HAADF-STEM image of the hetero-aggregate.
Finally, shot noise according to a typical electron dose of $\unit[149]{electrons/\text{\AA}^2}$ \cite{KRAUSE2016146} and scan noise according to a possible typical beam displacement of $\unit[0.01]{nm}$ \cite{10.1017/S1431927613001402} were applied.

\subsection{Statistical analysis and processing of simulated data}\label{sec.two.two}

In this section the need for the usage of neural networks is explained, addressing some problems connected with the reconstruction of preset values of the model  parameters 
$\fModel, \rModel, \cAModel,\cBModel$,    based on virtual aggregates drawn from    the stochastic 3D model. 
Further complications in this reconstruction task arise when using 2D STEM data instead of the full 3D geometry of the aggregates, through a loss of information. 
Therefore, it is explained how image processing methods can be used to simplify the extraction of information from STEM data.

\subsubsection{Estimating the parameters of the stochastic 3D model}\label{Data labeling}
       One of the challenges associated with predicting the  parameters 
        of the stochastic 3D model
       from STEM images is that some model parameters are even imperceivable from the 3D structure of a virtual aggregate  from which the corresponding STEM image  is determined. This is due to the simplifying assumptions made within the simulation process of the 3D model and its stochastic nature, see Section~\ref{section:Artificial aggregate generation},  
       resulting in empirical values of the model parameters slightly  differing from the preset ones.
       
    For example, the fractal dimension $\fRealized$ computed by means of Eq.~(\ref{Eq::FractalDimension}) for a virtual hetero-aggregate $A$ might differ from the preset value of the model parameter $\fModel$. More precisely, in the simulation of  hetero-aggregates,  the radii $r_1,\ldots,r_N$ of particles considered in Eq.~\eqref{Eq::FractalDimension} are replaced by their arithmetic mean  $(r_1+\ldots+r_N)/N$, see Section~\ref{section:Artificial aggregate generation}.
    Also, the mixing ratio $\rho$ of an aggregate $A$ computed by means of Eq.~\eqref{def.rho} can deviate from the model parameter $\rModel$, due to the randomly chosen labels of primary clusters, modeled by the Bernoulli-distributed random variables $L_1,\ldots,L_\numC$. For example, the first aggregate in Figure~\ref{figure:different_cluster_sizes} has an expected (preset) mixing ratio of $\rModel=0.3$, but the actual mixing ratio $\rho$ computed from Eq.~\eqref{def.rho} is $\rho=\frac{9}{41} \approx 0.22$. Recall that, in order to distinguish between these quantities, the vector of  model parameters used to generate $A$ is referred to as $\theta = ({\fModel}, {\rModel},{\cAModel},{\cBModel})$, while $\fRealized(A)$ and $\rRealized(A)$ describe the empirical fractal dimension and mixing ratio of the aggregate $A$, computed from Eqs.~(\ref{Eq::FractalDimension}) and \eqref{def.rho},  respectively.

\begin{figure}[ht]
            \centering
            \includegraphics[width=\textwidth]{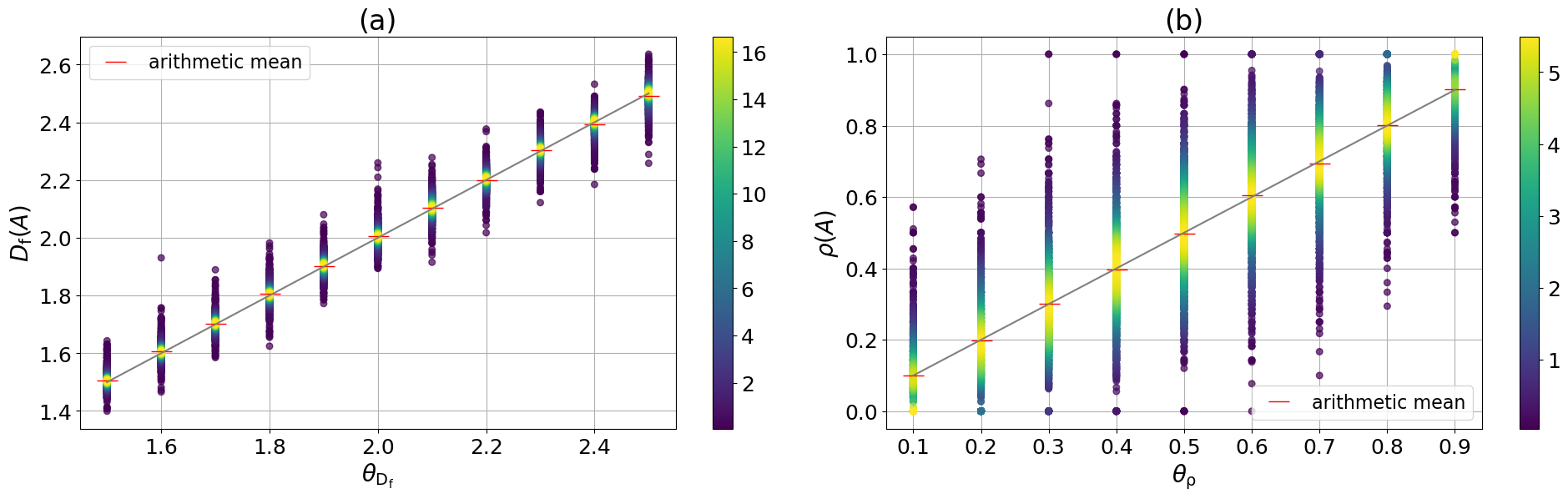}
            \caption{Visualization of empirical probability densities of the fractal dimension $\fRealized(A)$ (left) and mixing ratio $\rRealized(A)$ (right) of virtual hetero-aggregates, depending on the model parameters $\fModel$ and $\rModel$, where the values of $\fRealized(A)$ and $\rRealized(A)$ are computed by means of Eqs.~(\ref{Eq::FractalDimension}) and \eqref{def.rho},  respectively. The other model parameters were chosen at random from their respective ranges, as introduced in Section~\ref{section:Artificial aggregate generation}. To improve clarity, kernel density estimation was used to assign colors to the scatter points  computed for $19\,440$ realizations of the stochastic 3D model. }
            \label{verteilung von realisierten werten}
        \end{figure}

    Figure~\ref{verteilung von realisierten werten} illustrates the discrepancy between preset model parameters and the empirical fractal dimension and mixing ratio of  virtual aggregates, computed from Eqs.~(\ref{Eq::FractalDimension}) and \eqref{def.rho},  respectively.  Nevertheless, Figure~\ref{verteilung von realisierten werten} indicates that, on average, the structural descriptors $\fRealized$ and $\rRealized$ nicely coincide with the preset model parameters $\fModel$ and $\rModel$. Therefore, rather than attempting to determine the model parameters ${\fModel}, {\rModel},{\cAModel},{\cBModel}$
     from a single aggregate $A$, a family $B=\{A_1,\ldots,A_\nu\}$ of  $\batch>1$ aggregates is used instead, called \textit{batch} in the following. More specifically, it is expected that choosing a larger batch size would yield more accurate results, but at an increased cost.
    
    We were not able to find any scalar features that can be utilized to predict the model parameters $\cAModel$ and $\cBModel$ associated with the cluster size used in the generation of virtual aggregates, i.e., in the cluster-cluster-model introduced in Section~\ref{sub.sto.mod}.
    For example, in order to predict the model parameter $\cAModel$, an obvious choice for such a scalar feature would be to describe the average size of observable \ce{WO3} clusters, where an observable cluster is an inclusion maximal homogeneous subset $C\subset A$ of an aggregate $A$, i.e., there is no larger homogeneous subset $C'\subset A$ such that $C\subset C'$. These clusters can differ from the primary clusters used in the construction algorithm described in Section~\ref{sub.sto.mod}. Specifically, the observable clusters are formed by  unions of primary clusters, whereas, contrary to the latter ones, the observable clusters are recognizable in the 3D data, see Figure~\ref{figure:different_cluster_sizes} for a visualization.
    
\begin{figure}[ht]
    \centering
    \includegraphics[width=0.6\textwidth]{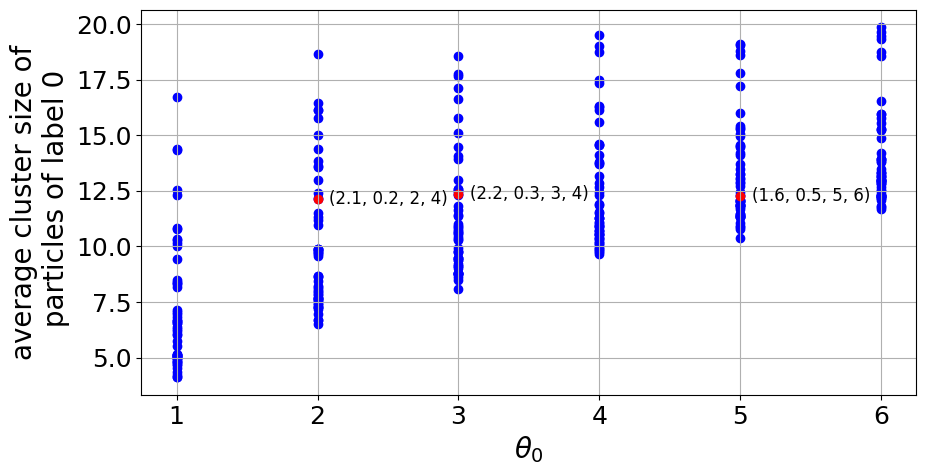}
    \caption{Average cluster size of \ce{WO3} particles. Each scatter point is computed over a batch $B=\{A_1,\ldots,A_\nu\}$  with $\nu=12$ for six different specifications of the model parameters $\fModel,\rModel,\cAModel,\cBModel$. For three of these specifications, the values of $\theta=(\fModel,\rModel,\cAModel,\cBModel)$ and the average cluster sizes, corresponding to the red scatter points, are displayed.}
    \label{challange_WO3}
\end{figure}

    However, this average (observable) cluster size can not be used to predict $\cAModel$. Figure~\ref{challange_WO3} shows that there are various specifications of model parameters that differ in $\cAModel$ and, nevertheless, yield similar average cluster sizes of \ce{WO3} particles.

    The prediction of the model parameter vector $\theta$ is further complicated by the fact that only the 2D STEM image can be utilized which may not perfectly inform the 3D morphology of $A$.

    To predict the preset vector of model parameters $\theta$ from a family $B$ of aggregates using only their simulated STEM images, CNNs are initially utilized to extract relevant features from these images. These features are subsequently utilized to predict the preset model parameters, see the schematic description of this workflow shown in Figure~\ref{figure:sketch_procedure}b. While the process of extracting features from the STEM images remains largely consistent across all model parameters, the calculation of the estimators for ${\fModel}, {\rModel},{\cAModel},{\cBModel}$ exhibits significant variations, see Sections~\ref{section:fractal_dimension}-\ref{section:Cluster size B} below.
    For instance, when estimating the model parameters $\fModel$ and $\rModel$, the features computed from a STEM image $I$ of an aggregate $A$ are scalar values that approximate $\fRealized(A)$ and $\rRealized(A)$. Then, the arithmetic mean of the respective image-wise features of a family $B=\{A_1,\ldots,A_\nu\}$ 
    of aggregates  is used as estimators $\fModelpredicted$ and $\rModelpredicted$ for $\fModel$ and $\rModel$. 
    In contrast, when predicting the model parameters $\cAModel$ and $\cBModel$, a neural network is employed to identify high-dimensional features from which the estimators $\cAModelpredicted$ and $\cBModelpredicted$ for $\cAModel$ and $\cBModel$ are computed, see Section~\ref{sec:cnn} below.\\

\subsubsection{Data processing and augmentation}\label{section:preprocessing}
        Various common image processing methods are used to simplify the extraction of information from STEM images. In particular, the pixel intensity values of  STEM images  are linearly scaled to the entire range of $[-0.5, 0.5]$ and rounded to $256$ equidistant values in order to achieve faster convergence to a lower error during the training process. More specifically, the scaling centers the pixel intensity values around zero~\cite{preprocess},  whereas the rounding reduces the noise of the images. Note that this procedure is performed on all STEM images, even if not explicitly mentioned, whereas the subsequent preprocessing steps will only be applied during training.
        
        Overfitting is a common problem where neural networks achieve good results on  training data but perform rather poorly when applied to previously unseen data. This  can occur when the model learns irrelevant information within the dataset. As a result, the model fits too closely to the training set and becomes overfitted, making it unable to generalize well to new data. To address this issue,  augmentation of training data is used. In the context of the present paper, this means that the input data is randomly modified in each training step, such that during each step of the training procedure the network is provided with input data which differs from the input data of previous steps. 
        Therefore, a significantly larger number of training steps can be conducted while still providing the neural network with novel training data in each step, and thus, avoiding overfitting.

        Note that there is a wide variety of possible methods for modifying input data which are commonly used in training data augmentation, e.g.,
        rotation, reflection, radial transformation, elastic distortion \cite{simard_best_2003} and random erasing \cite{10.1145/3510413}.
        However, in order to preserve certain structural descriptors of aggregates observed in image data,  like shape and size descriptors of particles, only random rotations, reflections and small displacements are used for training data augmentation. 

\subsection{CNN-based approach for the prediction of model parameters}
\label{sec:cnn}
The goal of this section is to introduce the CNN-based methodology for predicting the model parameters $ \fModel,\rModel, \cAModel, \cBModel$ of the hetero-aggregate model from  (simulated) STEM images.

Due to computational constraints, it was not feasible to generate the required number of aggregates for each possible preset of the model parameters $\fModel,\rModel, \cAModel, \cBModel$. Therefore, in order to ensure robust training, the focused was on generating 100 aggregates for each triple $(\rModel, \cAModel, \cBModel)$ in $\{0.1,\ldots, 0.9\} \times \{1,\ldots, 6\} \times \{1,\ldots, 6\}$, as these parameters exhibited interactive effects that were crucial for our study. More specifically, for each such triple, two values $\fModel^{(1)},\fModel^{(2)}$ of $\fModel$ were chosen at random from $\{1.5, \ldots, 2.5\}$ and each resulting model parameter preset $(\fModel^{(1)},\rModel, \cAModel, \cBModel),(\fModel^{(2)},\rModel, \cAModel, \cBModel)$ was used to generate 50 aggregates.
After applying the STEM simulation described in Section~\ref{sec.two.one}, this results in a set 
$G = \{(A_i,I_i,\theta_i) : 1 \leq i \leq 32\,400 \}
$ of 32\,400 triplets of 3D aggregates $A_i$, corresponding STEM images $I_i$, and vectors of preset model parameters $\theta_i = ({\fModel}_{,i}, {\rModel}_{,i},{\cAModel}_{,i},{\cBModel}_{,i})$. The set $G$ is thereafter split into two datasets, one for training and one for evaluation.

For both, training and evaluation, batches 
\begin{align}
B=\{(A_{i_1},I_{i_1},\theta_{i_1}),\ldots,(A_{i_v},I_{i_v},\theta_{i_v})\}\subset G
\label{def.bat.beh}
\end{align}
will be used for some $\nu>1$, that are generated by the same preset of model parameters, i.e., $\theta_{i_1}=\ldots=\theta_{i_v}$. To ensure the availability of such batches, the split of $G$ is done such that there is no model parameter configuration that occurs less than 20 times in neither the data used for training nor in the data used for evaluation. 
These two datasets will be referred to by their respective index sets $T$ (for training) and $E$ (for evaluation), where $T 
\cup E = \{1, \dots, 32\,400\}$ with $\#T=19\,440$ and $\#E=12\,960$. 

In the following, it is explained how the triplets $(A_i,I_i,\theta_i)$ are used to generate pairs of image data and ground truth labels, which will be utilized for the training of the neural networks. First, general aspects of network architecture and training are presented and, then,  some specifics regarding the prediction of each of the four model parameters 
${\fModel}, {\rModel},{\cAModel},{\cBModel}$
are given.

        \subsubsection{Network architecture and training}\label{section:architecture}
\begin{figure}[ht]
            \centering
            \includegraphics[width=11cm]{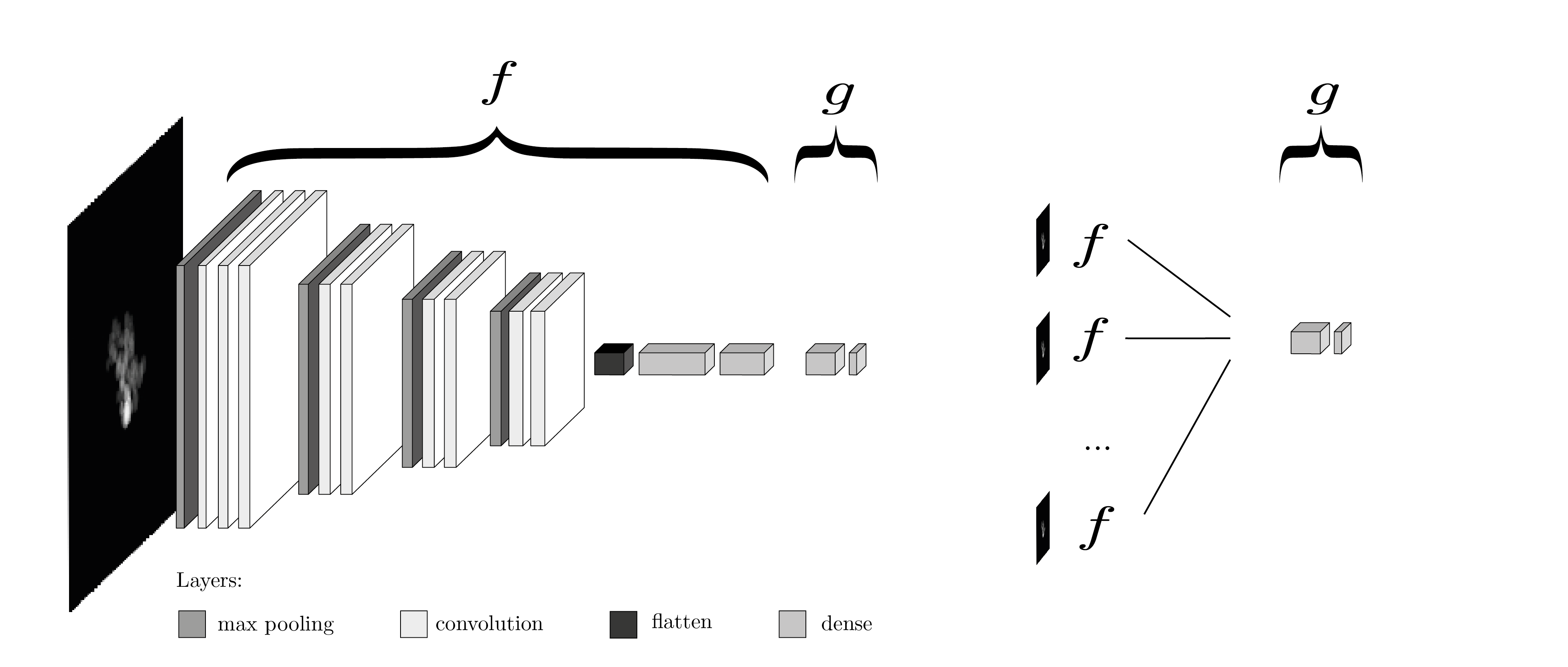}
            \caption{The basic network architecture represented as a simple composition of subnetworks $f$ and $g$ (left), and  an adjusted multi-image input network (right), where the subnetwork $f$ is applied to multiple input images and the outputs are  concatenated as a new input for subnetwork $g$ (see also Section \ref{Cluster size type A} below).}
\label{figure:architecture_image}
        \end{figure}

        The networks used to extract features are all based on the same basic network architecture, regardless of the model parameter being predicted. 
        This network architecture consists of stacked convolutional layers with a kernel size of $3\times3$, batch normalization layers~\cite{ioffe2015batch}, the ReLu activation function, given by
        \begin{align}
            \text{ReLu}(x)= \max\{0,x\}\qquad\mbox{for $x\in\R$,}
        \end{align}
        and max pooling layers with a kernel size of $2\times2$, followed by fully connected layers. The basic  architecture of the convolutional neural networks considered in the following  has the form 
        \begin{align}
            CNN = g(f)\,,
        \end{align}
        i.e., it is represented as  the composition of two subnetworks, $f$ and $g$. The subnetwork $f$ consists of the convolutional part of the basic network architecture, a flatten layer and two dense layers with a final output dimension of 112. The subnetwork $g$ consists of two dense layers with a final output dimension of 1. 
        A schematic representation of the network architecture is given in Figure~\ref{figure:architecture_image}~(left), whereas details regarding the this architecture are  provided in Table~\ref{architecture_table}.

        \begin{table}[ht]
        \footnotesize
        \begin{tabular}{l|l|c|c}

 & layer & output shape & number of parameters \\
\hline

0&input & $(512, 512, 1) \cdot b $ & $0      \cdot 1              $ \\
1&average pooling & $(256, 256, 1) \cdot b $& $0   \cdot 1         $ \\  
2&\convLayer & $(254, 254, 8) \cdot b $&   $(80 +  56 + 0) \cdot 1$ \\
3&\convLayer & $(252,252,16) \cdot b $& $(1\,168 + 112 + 0) \cdot 1$ \\
4&\convLayer & $(250,250,32) \cdot b $& $(4\,640 + 224 + 0) \cdot 1$ \\
5&\maxPoolLayer & $(125, 125, 32) \cdot b $& $0           \cdot 1$ \\ 

6&\convLayer & $(123, 123, 64) \cdot b $&   $(18\,496 +  448 + 0) \cdot 1$ \\           
7&\convLayer & $(121, 121, 64) \cdot b $&   $(36\,928 +  448 + 0) \cdot 1$ \\           
8&\maxPoolLayer & $(60, 60, 64) \cdot b $&      $0             \cdot 1$ \\

9&\convLayer & $(58, 58, 128) \cdot b $&   $(73\,856 +  896 + 0) \cdot 1$ \\            
10&\convLayer & $(56, 56, 128) \cdot b $&   $(147\,584 +  896 + 0) \cdot 1$ \\           
11&\maxPoolLayer & $(28, 28, 128) \cdot b $&      $0             \cdot 1$ \\  

12&\convLayer & $(26, 26, 256) \cdot b $&   $(295\,168 +  1\,792 + 0) \cdot 1$ \\         
13&\convLayer & $(24, 24, 256) \cdot b $&   $(590\,080 +  1\,792 + 0) \cdot 1$ \\          
14&\maxPoolLayer & $(12, 12, 256) \cdot b $&      $0             \cdot 1$ \\  

15&flatten      &   $(36\,864)  \cdot b $&           $0  \cdot 1      $ \\
16&dense + ReLu &       $(224) \cdot b  $&            $8\,257\,760 \cdot 1   	$ \\
17&dense + ReLu &           $(112) \cdot b   $&            $25\,200  \cdot 1  $ \\
18&dense + ReLu &       $(44)  $&            $112 \cdot b \cdot 44 + 44$ \\
19&dense &           $1   $&            $45 \cdot 1 $ \\
\hline 
&&&$= b \cdot 4\,928 + 9\,457\,713$ \\
\hline
        \end{tabular}
        \caption{Details of network architecture. The value of $b$ is put equal to $b=1$ for the prediction of the model parameters $\fModel$ and $\rModel$, and $b=\batch$ for the prediction of $\cBModel$ and $\cAModel$. Since padding is omitted, each convolutional layer reduces the size of the feature map by two in both dimensions.}
        \label{architecture_table}
        \end{table}

To achieve a high prediction quality, the parameters of the neural networks have to be adopted. This will be done supervised. More precisely, 
the dissimilarity between the
ground truth, denoted as  $y=(y_1,\ldots,y_n)$, and the 
 network output  $\widehat{y}=(\widehat{y}_1,\ldots,\widehat{y}_n)$, i.e.,
  $\widehat{y}_1=\mathrm{CNN}(x_1),\ldots,\widehat{y}_n=\mathrm{CNN}(x_n)$  for some input $x=(x_1,\ldots x_n)$, $n>1$, will be minimized. 
For example, when predicting the fractal dimension $\fModel$, the input $x$ of the network consists of STEM images $I_1,...,I_n$ and the ground truth  is given by the vector of fractal dimensions of the respective aggregates $A_1,...,A_n$, i.e. $y=(\fRealized(A_1),...,\fRealized(A_n))$. The comparison between the ground truth and the prediction is done in terms of the mean square error (MSE), given by
 \begin{align}
 \mathrm{MSE}(y,\widehat{y})= \frac{1}{n} \sum_{i=1}^{n}(y_i- \widehat{y}_{i})^2.\label{def.mse}
 \end{align}
The resulting loss $\mathrm{MSE}(y,\widehat{y})$ is  minimized by a gradient descent method using an Adam optimizer~\cite{Goodfellow2016,Kingma2015} with a learning rate of $0.0001$, where the value of $n$ in Eq.~\eqref{def.mse} determines the number of network evaluations before a step of the gradient descent method is applied. These evaluations are done on the training data, given by the index set $T$, where $n$ is put to $16$ when predicting $\fModel$ or $\rModel$, and $n=8$ otherwise.


The general network architecture described above and the prediction procedure will be slightly
adapted for  each of the four model parameters  $\fModel,\rModel,\cAModel$ and $\cBModel$.
In the following, detailed explanations will be provided regarding these  parameter-specific adaptations.

\subsubsection{Fractal dimension}
    \label{section:fractal_dimension}
    The fractal dimensions $\fRealized(A_{i_1}),\ldots,\fRealized(A_{i_\nu})$ of the aggregates $A_{i_1},\ldots,A_{i_\nu}$ in a batch $B$, as introduced in Eq.~\eqref{def.bat.beh},  are typically symmetrically distributed around the preset value of $\fModel$, which will be denoted by $\fModel(B)$ in the following, see Figure~\ref{verteilung von realisierten werten}a. Therefore, the  mean value $\fMean(B)$, given by 
    \[
    \fMean(B) = \frac{1}{\nu} \sum_{k=1}^\nu\fRealized(A_{i_k}),
    \]
     could be used as an estimator for $\fModel(B)$.  However,
     since the fractal dimensions $\fRealized(A_{i_1}),\ldots,\fRealized(A_{i_\nu})$   cannot be directly determined from the STEM images $I_{i_1},\ldots,I_{i_\nu}$, approximations $\fRealizedpredicted(I_{i_1}),\ldots,\fRealizedpredicted(I_{i_\nu})$ are used instead. These approximations are computed by a convolutional neural network $\CNNDf$, where the STEM images $I_{i_1},\ldots,I_{i_\nu}$ are used as input.
     Thus, finally, the estimator $\fModelpredicted(B)$ for $\fModel(B)$ is given by
   \begin{align}
        \fModelpredicted(B) = \frac{1}{\nu} \sum_{j=1}^\nu\fRealizedpredicted(I_{i_j}) = \frac{1}{\nu} \sum_{j=1}^\nu\CNNDf(I_{i_j}).\label{est.fra.dim}
   \end{align}
   The  architecture of the neural network $\CNNDf$ coincides with the one described in Section~\ref{section:architecture}. The activation function of the output layer is a scaled sigmoid function. This kind of activation function is a standard choice for NNs with bounded outputs. More precisely, the activation function is given by $\gamma(x)=\alpha \frac{1}{1 + e^{-x}} + \beta$ for $x\in\R$, where $\alpha=1.4$ and $\beta=1.3$ are selected to ensure that the network can represent the expected range of values for $\fRealized$, with added tolerances on each side of the expected range, see Figure~\ref{verteilung von realisierten werten}a. Note that the
    input of the network during training consists of augmented versions $a(I_i)$ of the STEM images $I_i$ for $i \in T$, i.e., images that arise from  $I_i$ by reflecting, rotating and displacing, as described in Section~\ref{section:preprocessing}. The corresponding supervisory signal consists of the fractal dimension of the corresponding aggregates. 
    Hence, the network training is conducted using pairs $(a(I_i), \fRealized(A_i))$, for $i\in T$.

\subsubsection{Mixing ratio}\label{section:mixing_ratio}
    In Figure~\ref{verteilung von realisierten werten}b the distribution of the mixing ratio of  aggregates in dependence of the model parameter  $\rModel$ is visualized. 
    From there, it is evident that the mixing ratios $\rRealized(A_{i_1}),\ldots,\rRealized(A_{i_\nu})$ of aggregates $A_{i_1},\ldots,A_{i_\nu}$ within a batch
    $B$, generated by the 3D model with a preset value of $\rModel(B)$, follow a distribution the mean of which is approximately equal to $\rModel(B)$.

    This suggests using a similar approach as described above in Section~\ref{section:fractal_dimension}.
    However, note that there are some aggregates with a mixing ratio $\rRealized(A_i)$ being equal to $0$ or $1$.
    A  neural network with an architecture as that of $\CNNDf$ does not reflect these discrete values properly.
    Therefore, the prediction procedure for the mixing ratio is slightly modified by initially classifying whether an image $I_i$ depicts an aggregate with mixing ratio of exactly $0$ or $1$, using a classification network $\CNNclass$, and afterwards predicting the mixing ratio of the corresponding aggregate, using a  regression network $\CNNreg$. 
    For this purpose, the  networks $\CNNclass$ and $\CNNreg$, having the same basic network architecture as described in Section~\ref{section:architecture} and a commonly used~\cite{activationFunctions} unscaled sigmoid function $\gamma(x)=\frac{1}{1 + e^{-x}}$ for $x \in \R$ as activation function in the output layer, are trained for the respective tasks.
    The training of the regression network $\CNNreg$ is done on pairs $(a(I_i),\rRealized(A_i))$, $i\in T$, of augmented STEM images and corresponding ground truth mixing ratios,
    whereas the training of the classification network $\CNNclass$  is done on pairs of augmented STEM images and corresponding binary class labels, where a class label of $0$ or $1$ identifies the corresponding aggregate as heterogeneous or homogeneous, respectively.
    
    However, it is a well-known problem that number-wise imbalanced classes can lead to poorly performing classifications since classifiers tend to neglect the underrepresented classes, also known as imbalance problem \cite{oversampling}. To address this issue, the augmented STEM images of homogeneous aggregates, which account for about 10\% of all images, were oversampled in the training procedure of the classifier to achieve balanced classes. 
    
    Finally, to predict the mixing ratio of an aggregate via its STEM image, the outputs of 
    $\CNNclass$, which identifies homogenous aggregates, and $\CNNreg$, which determines the mixing ratio, are combined. More specifically, for a STEM image $I$, the predicted mixing ratio $\rRealizedpredicted(I)$ of the corresponding aggregate is given by  
    \[
        \rRealizedpredicted(I) = 
            \begin{cases}
                \eta\left(\CNNreg(I)\right), & \text{if } \CNNclass(I)>0.5,  \\
                \CNNreg(I),            & \text{else,}
            \end{cases}
    \]
    where $\eta: [0,1]  \rightarrow \{0,1\}$ is the function that rounds a number $x\in[0,1]$ to its closest integer $\eta(x)\in\{0,1\}$. This results in the estimator $\rModelpredicted(B)$ for the preset model parameter $\rModel(B)$ of a batch $
B=\{(A_{i_1},I_{i_1},\theta_{i_1}),\ldots,(A_{i_v},I_{i_v},\theta_{i_v})\}$, given by 
    \begin{align}
\rModelpredicted(B) = \frac{1}{\nu} \sum_{j=1}^\nu\rRealizedpredicted(I_{i_j}).\label{est.mix.rat}
    \end{align}

    \subsubsection{Size of primary \ce{WO3} clusters}\label{Cluster size type A}
    In the procedures for predicting the model parameters $\fModel$ and $\rModel$, described above, the process of determining an estimator involved the identification of a scalar feature that describes an aggregate property, namely, the fractal dimension $\fRealized$ and the mixing ratio $\rRealized$, that is predominantly influenced by the corresponding model parameter. This scalar feature can be directly computed from the virtual 3D aggregates, and thus, it is possible to predict it  from the corresponding 2D STEM images. Consequently, using this scalar feature,  formulas for estimating the model parameter from this property has been derived, see Eqs.~\eqref{est.fra.dim} and \eqref{est.mix.rat}.

    Since the model parameter $\cAModel$ is designed to control the cluster sizes of \ce{WO3} particles  for the cluster-cluster-aggregation model introduced in Section~\ref{sub.sto.mod}, such a property should relate to the number of connected \ce{WO3} particles.
    However, the sizes of observable clusters are not only influenced by $\cAModel$ but also by $\rModel$. On the one hand, larger values of $\cAModel$ lead to larger primary cluster sizes of clusters of label 0 and thus larger observable clusters. Lower values of $\rModel$ lead to larger proportions of primary clusters of label 0. Therefore, it is more likely that two primary clusters that are in contact, share the material label 0, and thus, the expected size of observable clusters of label 0 increases, see Figure~\ref{challange_WO3}. 
    
    This makes the average of the observable cluster size on its own an unsuitable property for estimating the model parameter $\cAModel$.
    Therefore, one has to search for another feature that is functionally related to  $\cAModel$. Additionally, a functional relationship that suitably maps  features derived from STEM images to an estimator of $\cAModel$ may not be captured solely by an average, necessitating the search for another suitable function. However, these two steps can be quite complex and time-consuming if done heuristically.
    To address this, a data-driven approach utilizing a neural network is adopted. This approach allows us to determine the feature vectors and the formula that relates them to the corresponding model parameter $\cAModel$.
    More specifically, the identification of relevant features is conducted by means  of part $f$ of the basic network architecture described in Section~\ref{section:architecture}. 
    The subnetwork $f$ is applied to all images in a batch individually, and the concatenated results are then used as input of part $g$ of the basic network architecture, which is in charge of determining the relationship between the feature vectors determined by $f$ and the model parameter $\cAModel$.  In detail, this results in a modified network, denoted as CNN$^0$, 
    which is given by
    \begin{align}
{\rm CNN}^0(I(B)) = g ( f(I_{i_1}),\ldots,f(I_{i_\nu}) ), \label{eq:CNNCA}
    \end{align}
    where $I(B)=\{I_{i_1},\ldots,I_{i_\nu}\}$ denotes the STEM images corresponding to
    the aggregates $A_{i_1},\ldots,A_{i_\nu}$ in a batch $B$.
Referring to Table~\ref{architecture_table}, the feature vectors of STEM images up to the output of layer 17 are computed as before. Then these feature vectors of a batch are concatenated and used as input of layer 18.
    The final output layer uses a ReLu transfer function. The modified network architecture is illustrated on the right-hand side of  Figure~\ref{figure:architecture_image}.

Note that the approach described above differs from the commonly used technique where a network, denoted as $f^\prime$, takes multi-channel input data, i.e., in our case $f^\prime(I_{i_1},...,I_{i_\nu})$. Such an approach allows the network to detect spatially resolved interdependencies among the images. In contrast, our approach considered in Eq.~\ref{eq:CNNCA} employs identical CNNs $f$ for dimensionality reduction and feature extraction on each input channel individually. As a consequence, this  ensures uniform feature extraction for every input image while also reducing the number of trainable parameters in the CNN. The choice of this approach is rooted in the concept that each image within a batch  a priori contains the same information regarding the underlying model parameters, and the lack of spatial interdependence between the images which would be relevant for the prediction of model parameters.
    
    
    Due to the problem-specific architecture of the network $\CNNcA$, the training data no longer consists of pairs of individual images and corresponding ground truths. Instead, for each batch $B$, the training pair $(\{a(I_{i_1}),\ldots,a(I_{i_\nu})\},\cAModel(B))$ consists of a corresponding batch of augmented images and the underlying model parameter $\cAModel(B)$.
    
    Since the model parameter $\cAModel$ can only  take integer values, the output of the network has to be rounded to obtain a valid estimator for  $\cAModel$, given by $\cAModelpredicted(B)=\eta(\CNNcA(I(B))$, where  where $\eta: [0,\infty)  \rightarrow \{0,1,\ldots\}$ is the function that rounds a number $x\ge 0$ to its closest integer $\eta(x)\in\{0,1,\ldots\}$.

\begin{figure}[ht]
            \centering
            \includegraphics[width=.5\textwidth]{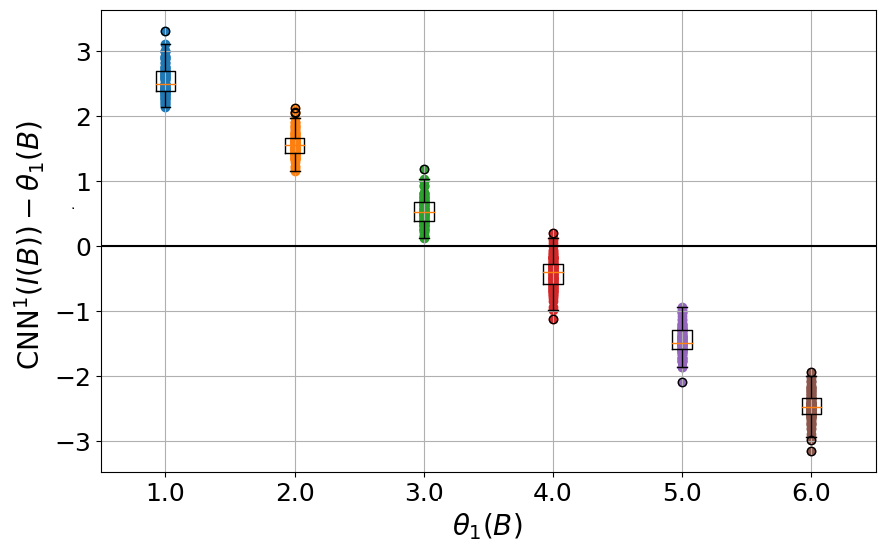}
            \caption{Prediction error of the network $\CNNcB$ for unmodified input images, in dependence of the preset value of the model parameter $\cBModel$. The prediction quality is comparable to that of a constant prediction, where $\CNNcB(I(B)) = 3.5$. The results have been obtained on a randomly chosen subset of the training data set.}
            \label{error_cB_invert=false}
        \end{figure}

    \subsubsection{Size of primary \ce{TiO2} clusters}\label{section:Cluster size B}

    The method used to predict the model parameter $\cBModel$ for the cluster size of \ce{TiO2} particles is similar to the approach described in the previous section.
    Nonetheless, given that the pixel intensity values of \ce{TiO2} particles in the STEM images closely resemble the background and are significantly lower than those of \ce{WO3} particles, they are considerably more difficult to differentiate by visual inspection. Consequently, it might be plausible that a neural network could also encounter challenges in tasks which depend on the identification of \ce{TiO2} particles. As shown in Figure~\ref{error_cB_invert=false}, the neural network  $\CNNcB$  achieves unsatisfactory results when using unadjusted image data, which may be due to the difficulty mentioned above.

\begin{figure}[ht]
            \centering
            \includegraphics[width=9cm]{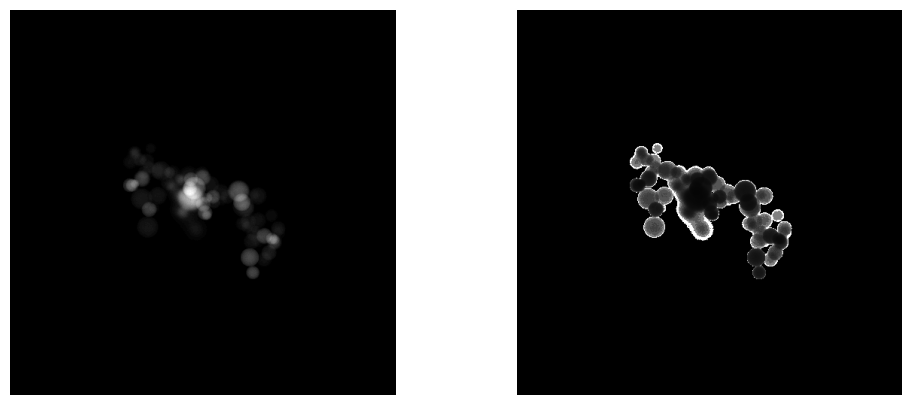}
            \caption{Effect of pixel intensity modification. The image on the right-hand side is obtained after replacing the intensity values 
             of  non-background pixels (shown on the left-hand side) by their multiplicative inverse, where a treshhold of $t=0.001$ is used.}
            \label{effect of invert}
        \end{figure}

    To address this issue, the intensity value $p>0$ of  non-background pixels in the STEM images is replaced by its multiplicative inverse $p^{\mathrm{modified}}$, i.e., for some threshold $t>0$ the modified pixel value is given by
       \[
        p^{\mathrm{modified}} = 
            \begin{cases}
                p, &  \quad\mbox{if $0<p<t$,} \\
               \displaystyle p^{-1},  & \quad \text{otherwise.}
            \end{cases}
    \]

    This procedure is applied to all STEM images used in the prediction of $\cBModel$ before the preprocessing steps described in Section~\ref{section:preprocessing} are applied. The highlighting effect of this adjustmemt of pixel intensity values is shown in Figure~\ref{effect of invert}.

\section{Results}\label{sec.three}

In this section, the results of the analysis on various aspects of model parameter prediction are presented. To ensure that these results accurately represent the generalization capability of the trained neural networks, all evaluations were conducted on data not used during training. More specifically, recall that the data corresponding to the index set $T$ is used for training, whereas the data corresponding to the index set $E$ is used to evaluate results, see Section~\ref{sec:cnn} for details on the training-test split. 

\begin{figure}[ht]
            \centering
            \includegraphics[width=\textwidth]{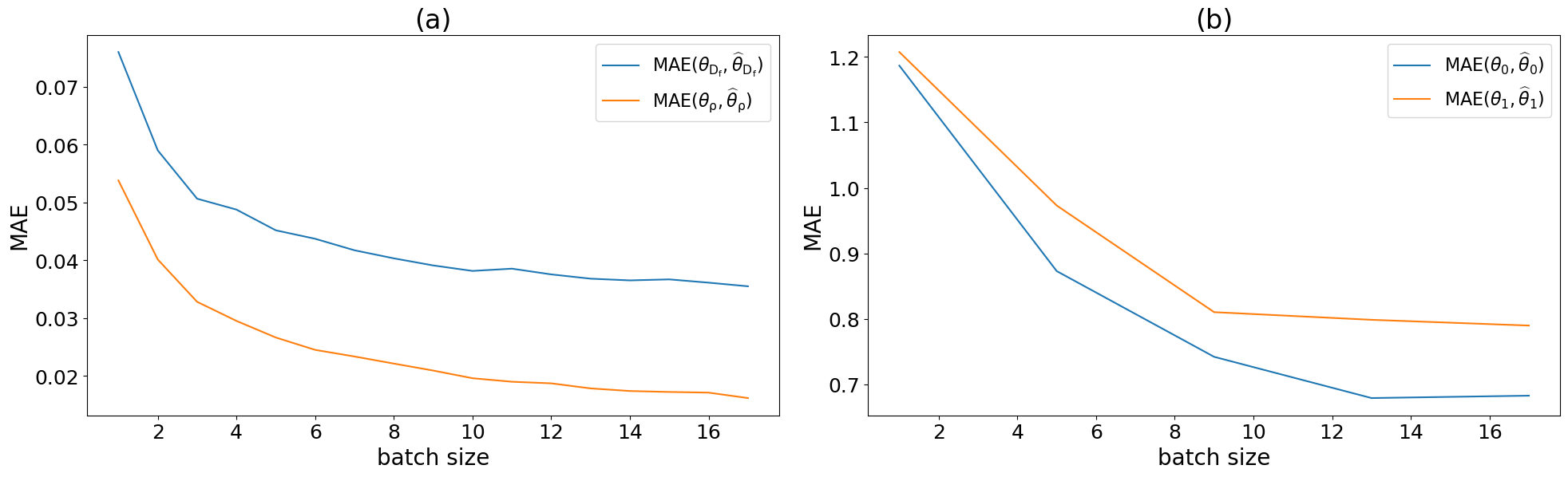}
            \caption{Quality of the estimators for 
${\fModel}, {\rModel},{\cAModel},{\cBModel}$, in dependence of the batch size $\batch$. Due to different orders of magnitude, the error curves for 
${\fModel}, {\rModel}$ and ${\cAModel},{\cBModel}$ are shown separately. The MAEs, defied by Eq.~\ref{def.mea.abs}, are computed over all available evaluation data, indexed by the set $E$.}
            \label{figure:batch_size_error}
        \end{figure}

As a prelude to the main findings, first, the impact of batch size on prediction quality is  assessed for all four model parameters 
${\fModel}, {\rModel},{\cAModel},{\cBModel}$.
For that purpose, Figure~\ref{figure:batch_size_error} illustrates how the batch size affects the quality of the predictions with respect to the mean absolute error (MAE), defined as
\begin{align}\mathrm{MAE}(y,\widehat{y})= \frac{1}{n} \sum_{i=1}^{n}|y_i- \widehat{y}_{i}|,\label{def.mea.abs}
\end{align}
where $n>0$ is the number of predictions and $\widehat{y}=(\widehat{y}_i)_{i=1, ..., n}$ are the predictions of the ground truth values $y=(y_i)_{i=1, ..., n}$. Note that the mean absolute error given in Eq.~\eqref{def.mea.abs} is more robust to outliers and yields  more easily interpretable values compared to the mean squared error considered in Section~\ref{sec:cnn}. As expected, it can be observed that larger batch sizes lead to better predictions. However, no significant improvement is observed for values exceeding 10. Thus, the results presented below, which were computed with a fixed batch size of $\nu=12$, can be considered representative for the presented methodology.

\subsection{Fractal dimension}
 The accuracy of the estimator $\fModelpredicted$ for $\fModel$ depends on two key properties. First, the mean error of the single STEM image predictions $\CNNDf(I)$ should be centered around zero, since otherwise a bias could be propagated through the averaging procedure and therefore bias the estimator $\fModelpredicted$, see Eq.~\eqref{est.fra.dim}. Second, the variance of the single image prediction error should be low, so a low variance estimator can be achieved even with a small batch size $\batch$.
 
\begin{figure}[ht]
        \centering
        \includegraphics[width=\textwidth]{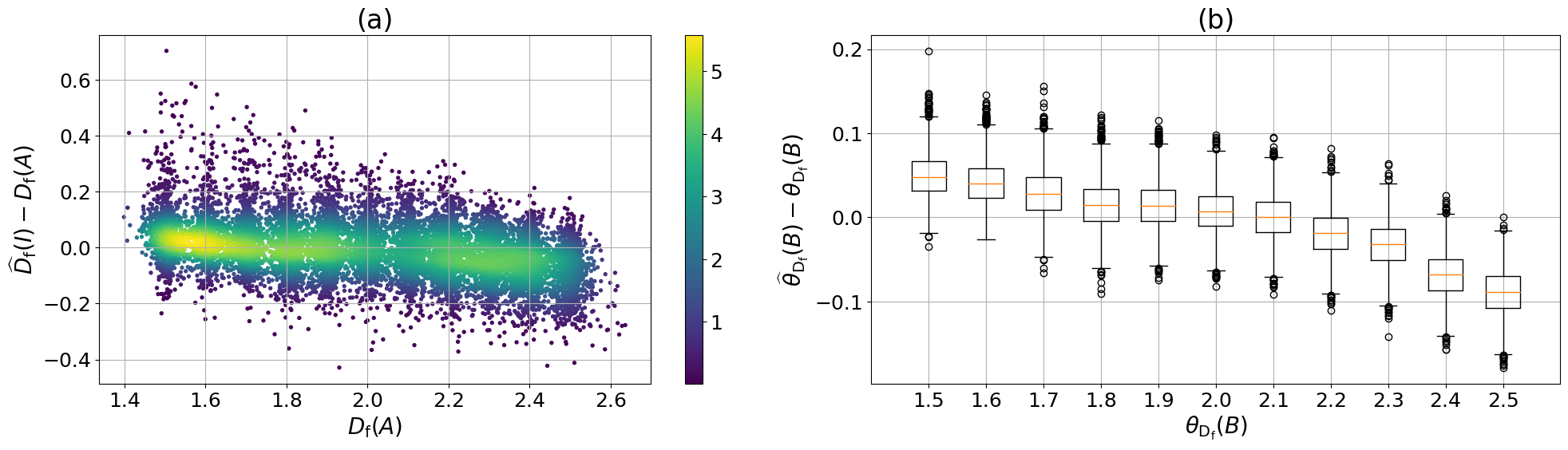}
        \caption{Estimation error of $\fRealizedpredicted$ (left) and $\fModelpredicted$ (right). In Subfigure (a) the quality of the prediction of the fractal dimension per aggregate is visualized, where the colors are computed by means of a Gaussian kernel density estimator. In Subfigure (b), the error regarding the prediction of the  model parameter $\fModel$ is shown, where  batches of size $\batch = 12$ are used for the computation of $\fModelpredicted$.}
        \label{result_fractal_dimension}
\end{figure}

In Figure~\ref{result_fractal_dimension}a the error for the predicted fractal dimension $\fRealizedpredicted$ is shown. As desired, the error of the network output exhibits a small absolute value for the bias and a low variance, as indicated by a mean value of -0.006 and interquartile range of 0.118. 
As the network output is a suitable basis for 
predicting the   model parameter $\fModel$,  the estimator $\fModelpredicted$ achieves an MAE of 0.041, see Figure~\ref{result_fractal_dimension}b. The network tends to slightly overestimate the fractal dimension of the depicted aggregates for small preset values of $\fModel$ and underestimate it for large ones. This behavior is further pronounced in the estimator $\fModelpredicted$. For a possible explanation of this trend, see Section~\ref{sec.four} below.


\subsection{Mixing ratio}\label{sec:result_mixing}
\begin{figure}[ht]
    \centering
    \includegraphics[width=\textwidth]{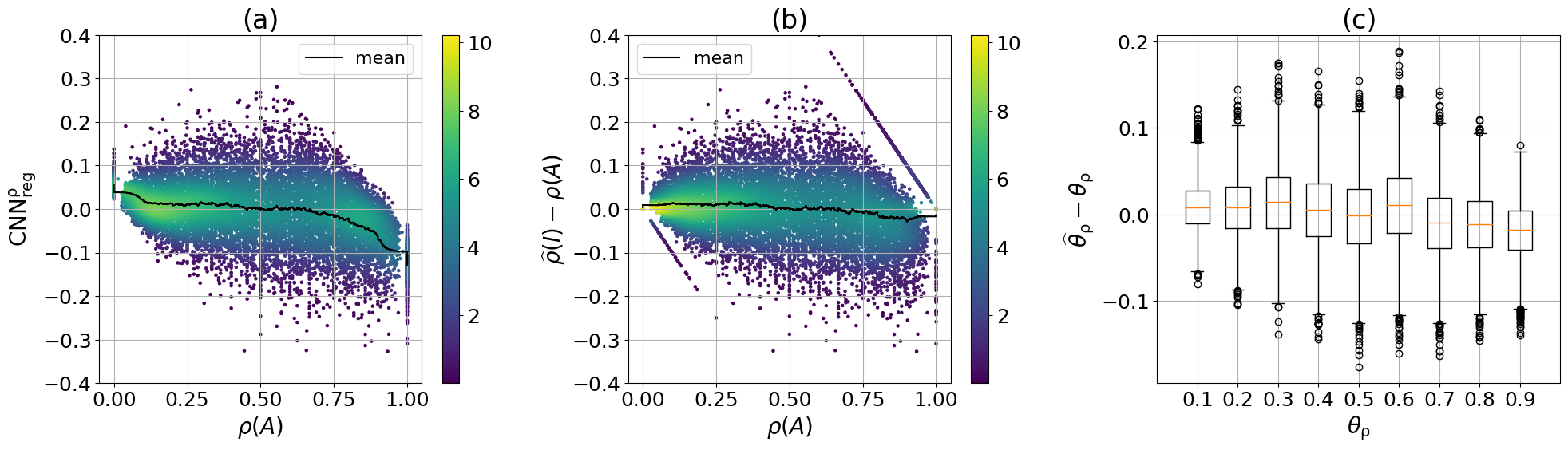}
    \caption{Estimation error of $\rRealizedpredicted$ (left, center) and $\rModelpredicted$ (right). In Subfigure~(a) the error of the unrounded output of the regression network is displayed per image. The black line, visualizing the mean error, is computed via a sliding window. The error for the  output of the modified network is shown in Subfigure~(b). In Subfigure~(c) the error regarding the prediction  of the model parameter $\rModel$ is displayed, where  the modified values of $\rRealizedpredicted$ are used.}
    \label{result_mixing_ratio}
\end{figure}

To evaluate the accuracy of the estimator for the model parameter $\rModel$, first, the image-wise straightforward case is considered, where the output  $\CNNreg(I)$ of the regression network is used as an estimator for the mixing ratio $\rRealized(A)$, without considering the classification network.
As shown in Figure~\ref{result_mixing_ratio}a, the  output  $\CNNreg(I)$ of the regression network exhibits a relatively high bias for aggregates $A$  such that $\rRealized(A) \in [0,0.1]$ or $\rRealized(A) \in [0.9,1]$, with biases of about  $0.04$ and $-0.1$, respectively.

To address this issue, in Section~\ref{section:mixing_ratio} a procedure which utilizes an additional classification network $\CNNclass$ is presented. In Figure~\ref{result_mixing_ratio}b, the resulting image-wise error of $\rRealizedpredicted$ using this procedure is shown. It is evident that the  error of $\rRealizedpredicted$ is significantly reduced for homogenous aggregates. More precisely, the bias of $\rRealizedpredicted(I)$ for aggregates $A$ with $\rRealized(A) \in [0,0.1]$ or $\rRealized(A) \in [0.9,1]$ decreases to about 0.009 and 0.02, respectively. Incorporating the additional network, the MAE of the image-wise predicted mixing ratio $\rRealizedpredicted(I)$ of an aggregate $A$ decreases from 0.059 to 0.053. Consequently, the MAE of the batch-wise prediction $\rModelpredicted$ of $\rModel$ improves significantly, reducing from $0.027$ to $0.017$.

Note that the 
diagonally arranged points in Figure~\ref{result_mixing_ratio}b are due to a small number of falsely classified heterogeneous aggregates, whereas the significantly thinned vertical lines are due to correctly classified homogeneous aggregates. The amounts of correctly and falsely classified aggregates are displayed in Table~\ref{table:confusion_matrix}.

\begin{table}[ht]
\centering
\begin{tabular}{|l|c|c|} 
\hline
prediction~\textbackslash~true & homogenous & heterogeneous \\
\hline
homogenous & 1\,174 & 567 \\
\hline
heterogeneous & 67 & 12\,082 \\
 \hline
 \end{tabular}
\caption{Confusion matrix of the homogeneous-heterogeneous classification task. About 95\% of the aggregates are correctly classified.}
\label{table:confusion_matrix}
\end{table}

\subsection{Sizes of primary \ce{WO3} clusters and primary  \ce{TiO2} clusters}
\begin{figure}[ht]
    \centering
    \includegraphics[width=\textwidth]{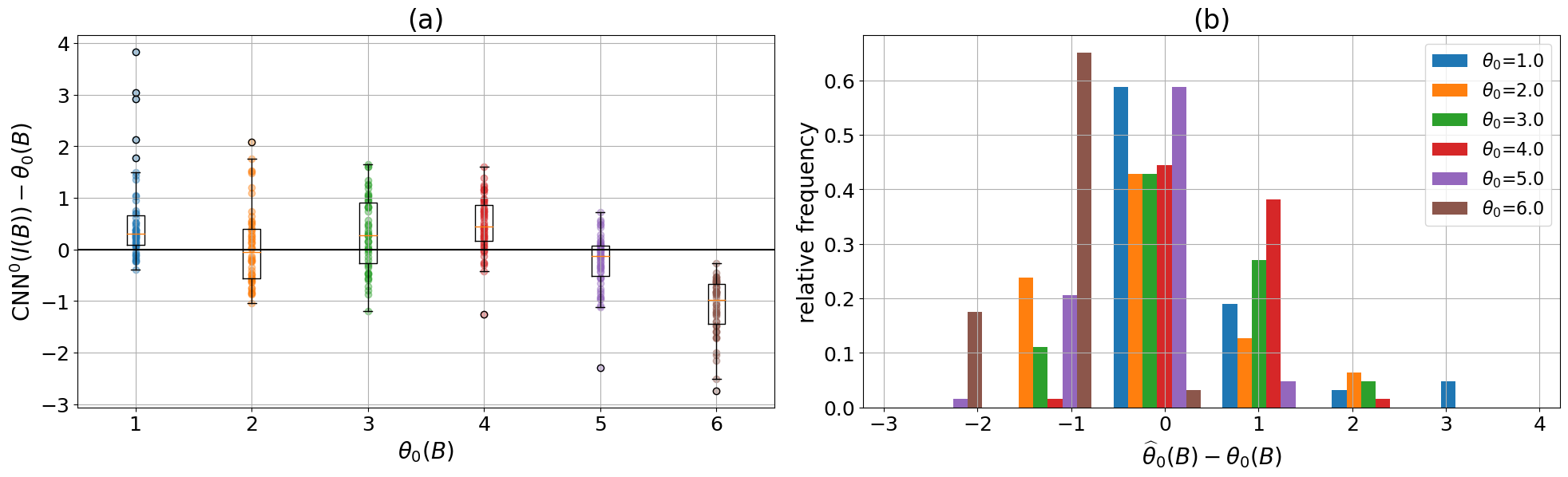}
    \caption{Estimation error of $\cAModelpredicted$. In Subfigure~(a), the  differences between the network output  $\CNNcA(I(B))$
   and the preset values $\cAModel(B)$ of the model parameter $\cAModel$ are shown. The prediction error of $\cAModelpredicted$ after applying the rounding operation is displayed in Subfigure~(b).}
    \label{figure:error_cA}
\end{figure}

Figure~\ref{figure:error_cA}a shows the difference between the network output  $\CNNcA(I(B))$ 
for the STEM images  $I(B)=\{I_{i_1},\ldots,I_{i_\nu}\}$  corresponding to
    the aggregates $A_{i_1},\ldots,A_{i_\nu}$ in a batch $B$  (given in Eq.~\eqref{eq:CNNCA}, i.e., prior to rounding of the output which would result in the estimator $\cAModelpredicted$) and the preset value $\cAModel(B)$ of the model parameter $\cAModel$. Figure~\ref{figure:error_cA}b shows the error distribution of $\cAModelpredicted$ after rounding, where in about 48\% of all  cases the value of $\cAModelpredicted$ coincides with $\cAModel$. Additionally, in more than 92\% of the cases, the error of $\cAModelpredicted$ is less than or equal to 1. 
Although the largest mean absolute error occurs in the case of $\cAModel=6$, the resulting inaccuracy corresponds to an average relative error of about $20\%$.

\begin{figure}[ht]
    \centering
    \includegraphics[width=\textwidth]{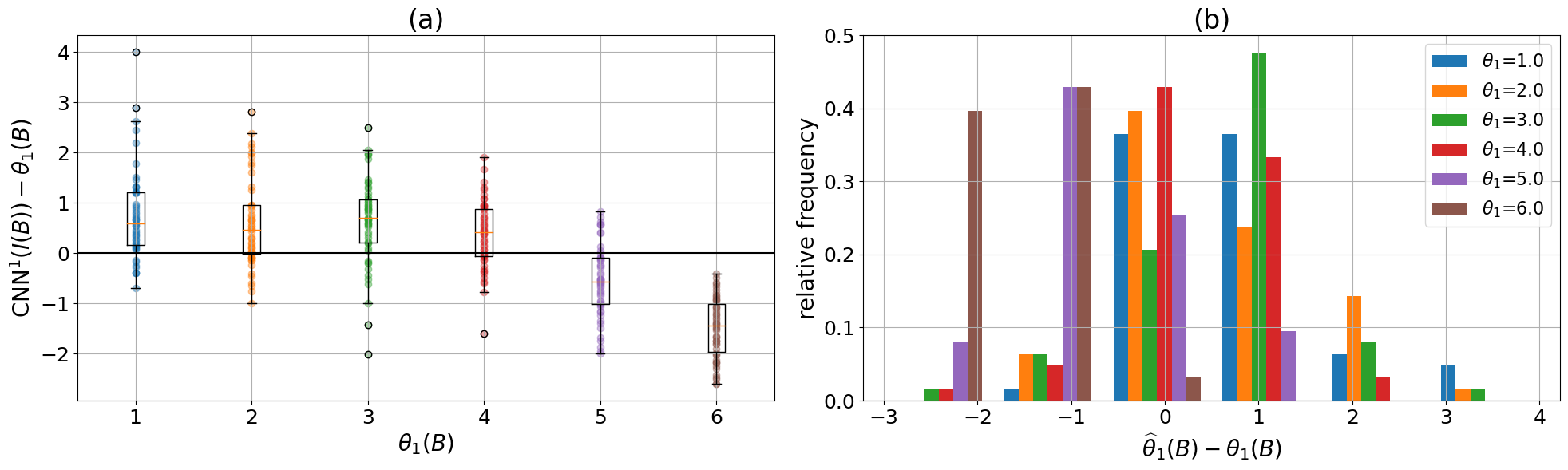}
    \caption{Estimation error of $\cBModelpredicted$. In Subfigure~(a), the  differences between the network output  $\CNNcB(I(B))$
   and the preset values $\cBModel(B)$ of the model parameter $\cBModel$ are shown. The prediction error of $\cBModelpredicted$ after applying the rounding operation is displayed in Subfigure~(b). }
    \label{figure:error_cB}
\end{figure}

The quality of the estimator $\cBModelpredicted$ 
introduced in Section~\ref{section:Cluster size B} is similar to that of
$\cAModelpredicted$, see Figure~\ref{figure:error_cB}. After rounding the output of the network $\CNNcB$,  32\% of the predictions coincided with the preset values of $\cBModel$. In about 82\% of the cases, an error less than or equal to 1 occurred. The mean absolute error for $\cBModel = 6$ is equal to $1.44$, where the resulting inaccuracy corresponds to an average relative error of about $24\%$.

\subsection{Further  structural descriptors of hetero-aggregates}\label{sec::structural_descriptors}

Recall that the goal of the method presented in this paper is to generate realistic digital shadows of hetero-aggregates in 3D, solely from observations provided by 2D STEM images of the aggregates. For that purpose, the parameters 
${\fModel}, {\rModel},{\cAModel},{\cBModel}$
of the stochastic 3D model introduced in Section~\ref{sub.sto.mod} 
are predicted in order to specify the model configuration with which to generate digital shadows. However, so far, only the accuracy of the predictors 
${\fModelpredicted}, {\rModelpredicted},{\cAModelpredicted},{\cBModelpredicted}$ for
${\fModel}, {\rModel},{\cAModel},{\cBModel}$ was evaluated, rather than investigating further structural descriptors of hetero-agggregates in order to evaluate the structural similarity between the resulting digital shadows and the original hetero-aggregates, i.e., the aggregates which were used for predicting the model parameters 
${\fModel}, {\rModel},{\cAModel},{\cBModel}$. Moreover, many structural properties of the digital shadows are influenced by multiple model parameters, and thus, evaluating the quality of the four predictors 
${\fModelpredicted}, {\rModelpredicted},{\cAModelpredicted},{\cBModelpredicted}$  separately is not sufficient. Therefore, three further structural descriptors, which characterize the 3D morphology of hetero-aggregates and have not yet been considered in this paper, are investigated in order to assess the similarity between original aggregates and corresponding digital shadows, see also Figure~\ref{figure:sketch_procedure}c-d. 

\subsubsection{Average cluster size and  coordination numbers}
\label{sec.ave.clu}

The average cluster size $\finalClusterSizeB(A)$
of \ce{TiO2} particles of an aggregate  
$
A= \{p_i=(x_i,r_i,l_i):x_i \in \R^3, ~ r_i \in\R^+,~ l_i \in  \{0,1\},~ 1\leq i\leq N \}.
$
describes the average cardinality of clusters of connected \ce{TiO2} particles in $A$. It is given by 
\begin{align}
    \finalClusterSizeB(A) = \frac{1}{\#C_{\ce{TiO2}}(A)} \sum_{c \in C_{\ce{TiO2}}(A)} \#c,
\end{align}
where $C_{\ce{TiO2}}(A)$ denotes the set of all \ce{TiO2} clusters in $A$. 
While the value of $\finalClusterSizeB(A)$ is primarily influenced by the preset values of 
$\cBModel$ and $\rModel$, the value of $\cAModel$ also has some (minor) influence on $\finalClusterSizeB(A)$  through its appearance in the definition  of the  Bernoulli-distributed  labels  $L_k$ of
the stochastic 3D model, see Section~\ref{sub.sto.mod}.

Furthermore, the so-called average hetero-coordination number $
\coordinationHetero(A)$ of an aggregate $A$ is considered, which is given by 
\begin{align}
\coordinationHetero(A)&=\frac{1}{\text{\#$A$}} \sum_{p\in A}
\# \{p^\prime \in A \colon p, p^\prime \text{ are in contact}, l \not= l^\prime \}\nonumber
\\
&=\frac{2 \text{\#\{set of heterogeneous contacts in $A$\}}}{\text{\#$A$}}\,,
\end{align}
where \#$A (=N)$
is the total number of particles in $A$. Thus, 
$
\coordinationHetero(A)$  is the average number of contacts of  particles in $A$  with  particles of the other material. Finally, the average coordination number $\coordinationTotal(A)$, given by
\begin{align}
\coordinationTotal(A)&=\frac{1}{\text{\#$A$}} \sum_{p\in A}
\# \{p^\prime \in A \colon p, p^\prime \text{ are in contact} \}\nonumber
\\ &=\frac{2 \text{\#\{set of contacts in $A$\}}}{\text{\#$A$}}\,,\label{def.tot.coo}
\end{align}
is considered, which is the average number of contacts of  particles in $A$  to other particles, regardless of their material.

Since the number of contacts of a particle within an aggregate $A$ strongly depends on the  shape of $A$, the model parameter $\fModel$ significantly influences the values of the descriptors $\coordinationHetero(A)$ and $\coordinationTotal(A)$. Further,  $\coordinationHetero(A)$ tends to increase with $\rModel$ close to 0.5 and decreasing primary cluster sizes determined by $\cAModel$ and $\cBModel$.

\subsubsection{Comparison of original hetero-aggregates and their digital shadows}\label{sec.com.ori}

\begin{figure}[ht]
    \centering
    \includegraphics[width=\textwidth]{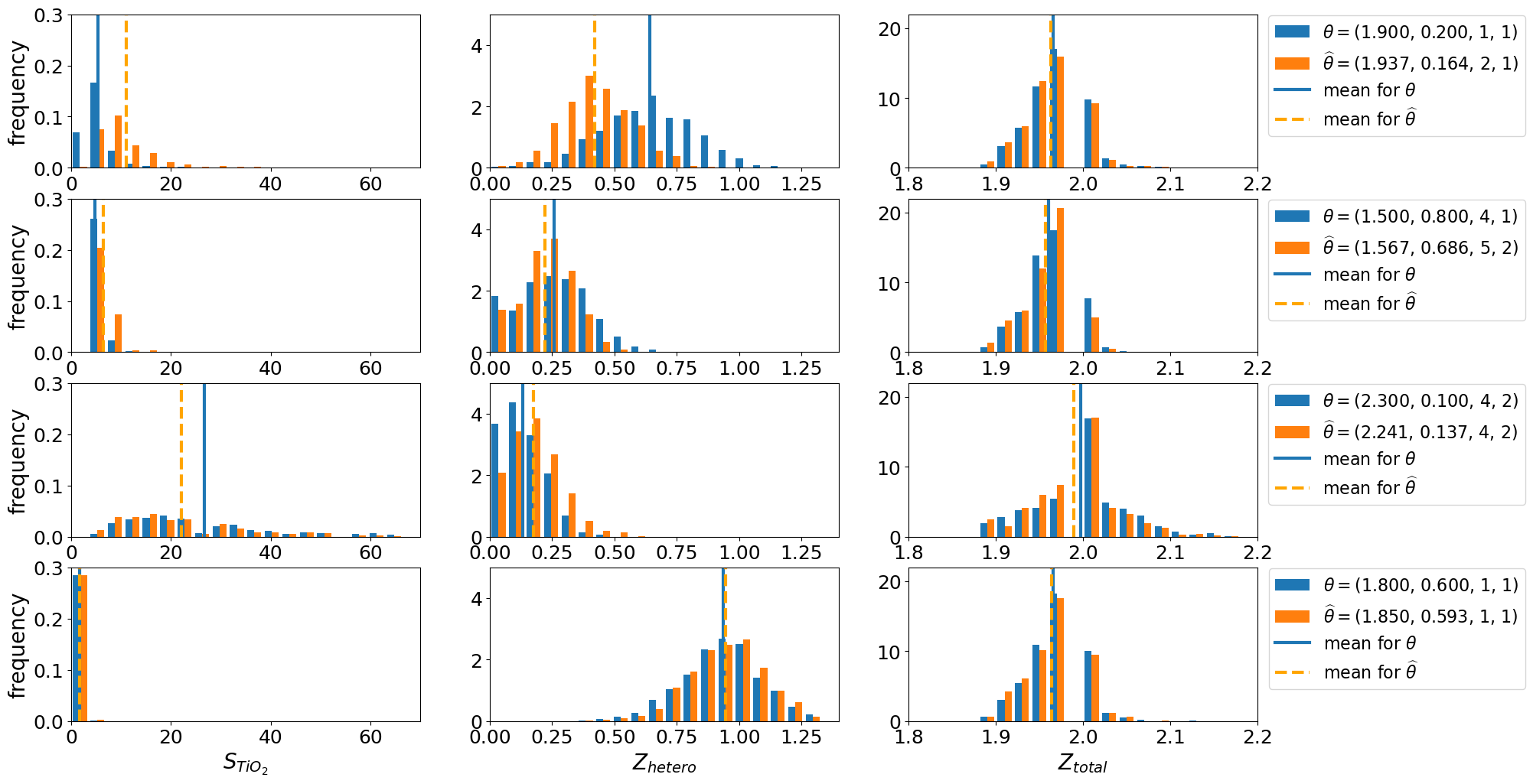}
    \caption{Distribution of the structural descriptors $\finalClusterSizeB(A_i)$ and $\finalClusterSizeB(A_i^\prime)$ (left column),  
$\coordinationHetero(A_i)$ and $\coordinationHetero(A_i^\prime)$ (middle column), as well as
$\coordinationTotal(A_i)$ and  $\coordinationTotal(A_i^\prime)$ (right column) of the original aggregates $A_i$ (blue) and their digital shadows $A_i^\prime$ (orange), for four numerical specifications of $\theta$ and their  predictions $\widehat\theta$.
For computing the histograms,  20 equidistant bins have been employed which span the entire range of respective values on the $x$-axis.}
\label{figure:coordination_avgB}
\end{figure}

To evaluate the quality of the predictor $\widehat{\theta}=(\fModelpredicted,\rModelpredicted,\cAModelpredicted,\cBModelpredicted)$ in terms of the structural descriptors introduced in Section~\ref{sec.ave.clu}, 50  configurations of $\theta = (\fModel,\rModel,\cAModel,\cBModel)$ were selected at random, out of the index set $E$ of
 evaluation data. For each of these numerical specifications of $\theta$, 800 new aggregates $A_1,\ldots,A_{800}$  were drawn from the corresponding stochastic 3D model, and their  structural descriptors $\finalClusterSizeB(A_i), \coordinationHetero(A_i)$ and $\coordinationTotal(A_i)$ for $i\in\{1,\ldots,800\}$ were computed.
Furthermore, for each case,  the (preset) ground-truth parameter vector $\theta$ has been estimated using the methods explained in Section~\ref{sec:cnn}.  Then, for each of the 50 specifications of $\widehat{\theta}$, 800 additional aggregates 
$A_1^\prime,\ldots,A_{800}^\prime$ 
and computed their structural descriptors 
$\finalClusterSizeB(A_i^\prime), \coordinationHetero(A_i^\prime)$ and $\coordinationTotal(A_i^\prime)$ for $i\in\{1,\ldots,800\}$ were generated.
Figure~\ref{figure:coordination_avgB} visualizes the distributions of these structural descriptors for four numerical specifications of $\theta$, where
the aggregates 
$A_1,\ldots,A_{800}$ 
and 
$A_1^\prime,\ldots,A_{800}^\prime$ 
were generated using either the preset  parameter vector $\theta$ (blue) or its prediction $\widehat{\theta}$ (orange), respectively.

Note that the gaps in the histograms of the average coordination numbers 
$\coordinationTotal(A_i)$ and  $\coordinationTotal(A_i^\prime)$ (right column) 
are due to the limited size of the considered aggregates, see Section~\ref{section:Artificial aggregate generation}. More specifically, 
the average coordination numbers 
$\coordinationTotal(A_i)$ and  $\coordinationTotal(A_i^\prime)$  given in Eq.~\eqref{def.tot.coo}, of aggregates $A_i,A_i^\prime$ with  sizes smaller than or equal to $80$,
 can only take values in the set 
 \begin{align}
H=\{\frac{2q_1}{q_2}:q_1,q_2\in \N, q_1 \leq q_2 \leq 80\}, \label{def.set.hah}
\end{align}
where $H \cap (1.975,2) = \emptyset$ because of the limited denominator $q_2$ on the right-hand side of Eq.~\eqref{def.set.hah}. 

Furthermore,  note that the predictor $\widehat{\theta}$ for  $\theta$ displayed in the top row of Figure~\ref{figure:coordination_avgB}  has a much smaller mean absolute error than the one displayed in the second row. Nevertheless, the latter (blue and orange)
histograms show a higher agreement than those in the top row of Figure~\ref{figure:coordination_avgB}.
Meaning that a high degree of similarity (in terms of MAE) of $\theta$ and $\widehat\theta$ does not necessarily imply a high degree of similarity of the resulting descriptor distributions.

We  quantitatively analyzed this discrepancy between the distributions of the structural aggregate descriptors resulting from the preset configuration of model parameters  and their prediction. For that purpose, the absolute difference  of the means of these pairs of distributions were computed . For example, the mean values of $\finalClusterSizeB(A_i)$ and $\finalClusterSizeB(A_i^\prime)$ (vertical lines)  in the top row of Figure~\ref{figure:coordination_avgB} are equal to $5.38$ and $ 11.00$ for the preset parameter vector $\theta$ and its prediction $\widehat{\theta}$, respectively. This results in an absolute error of  $5.62$. Over all 50 pairs  of $\theta$ and  $\widehat\theta$, a MAE error of $2.165$ is achieved, see also
Table~\ref{Tab::error}, where  the  MAEs for all three  structural descriptors considered in this section are given as well as the corresponding coefficient of determination $R^2$
defined as
\[R^2(y,\widehat{y})= 1-\frac{\mathrm{MSE}(y,\widehat{y})}{\mathrm{MSE}(y,\overline{y})}.\]
 Here, the vectors $y=(y_1,\ldots,y_{50}),\widehat{y}=(\widehat y_1,\ldots,\widehat y_{50})\in \R^{50}$ consist of the mean values of the distributions of the given aggregate descriptor  computed for the 50  preset specifications of $\theta$ and their predictions $\widehat\theta$. 
 More precisely, for $j\in\{1,\ldots,50\}$,  $y_j =\frac{1}{800} \sum_{i=1}^{800}\gamma(A_{ij})$ and $\widehat{y}_j = \frac{1}{800}\sum_{i=1}^{800}\gamma(A^\prime_{ij})$, where $\gamma$ stands for either $\finalClusterSizeB,\coordinationHetero$ or $\coordinationTotal$, and $A_{ij},A^\prime_{ij}$ denote  the $i$-th aggregate drawn from the $j$-th specification of 
 $\theta$ and its prediction $\widehat\theta$,  respectively.
 Furthermore, $\overline{y} = \frac{1}{50} \sum_{i=1}^{50}y_i$. 


 \begin{table}[ht]
    \centering
    \renewcommand{\arraystretch}{1.5}
    \begin{tabular}{l|c|c|c} 
        \hline
         & $\finalClusterSizeB$ & $\coordinationHetero$ & $\coordinationTotal$ \\
        \hline
        MAE & 2.165 & 0.056 & 0.007 \\
        $R^2$ & 0.84 & 0.85 & 0.87 \\
        \hline
    \end{tabular}
    \renewcommand{\arraystretch}{1}
    \caption{Discrepancy between the mean values of the distributions 
     of  $\finalClusterSizeB$, $\coordinationHetero$, and $\coordinationTotal$, with respect to the mean absolute error MAE and the coefficient of determination $R^2$, computed for the 50  preset specifications of $\theta$ and their predictions $\widehat\theta$.  }
    \label{Tab::error}
\end{table}

\section{Discussion}\label{sec.four}

\begin{figure}[ht]
    \centering
    \includegraphics[width=\textwidth]{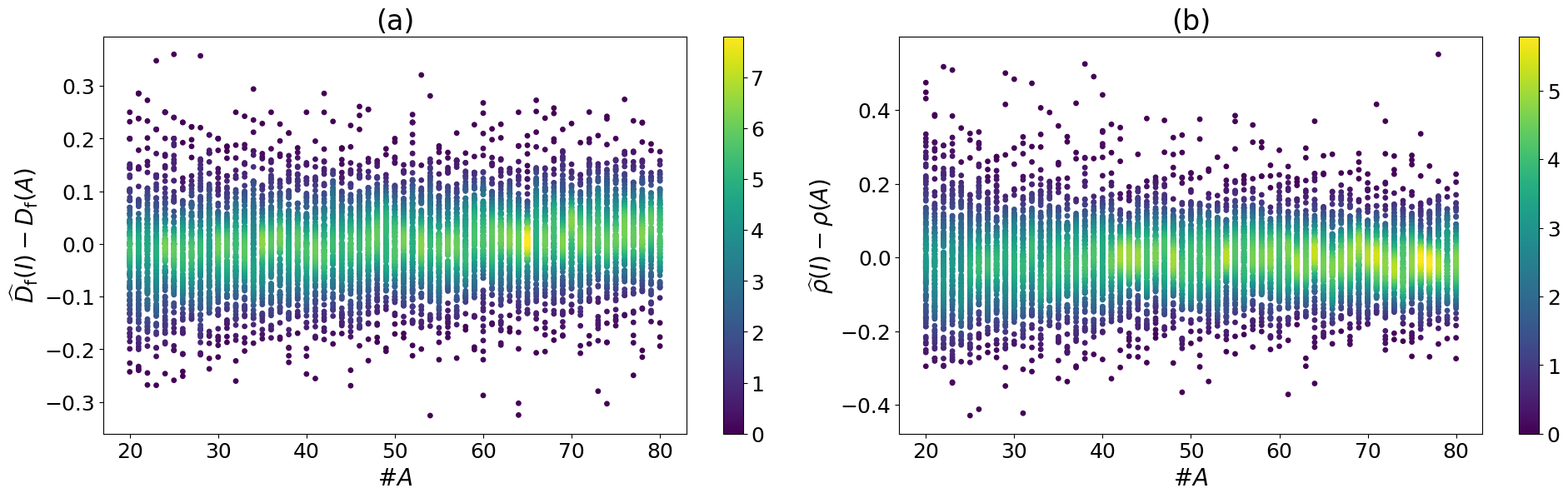}
    \caption{Prediction error for fractal dimension (left) and mixing ratio (right),  depending on the number of particles per aggregate, for aggregates taken from  the evaluation data
    given by the index set $E$. The color of  dots  is chosen according to a 1D Gaussian kernel density estimator along the $y$-axis.
    }
    \label{effect_aggregate_size}
\end{figure}

The analysis of image data in order to determine the fractal dimension of finite aggregates has  been a popular approach for some time. Two commonly used methods for this purpose are the box counting and sandbox methods, which are relatively simple image analysis tools~\cite{BUSHELL20021}. These methods can provide meaningful structural information, but the quality of the results is highly dependent on the quality of the images. Specifically, high contrast and resolution are necessary to obtain clear STEM images from which accurate structural information can be extracted.
However, in cases where a high fractal dimension is present, i.e., $\fRealized(A)>2$, these classical methods have to be adopted to avoid problems with geometric opacity. There are attempts to solve this problem under certain conditions, see~\cite{PhysRevE.69.011405}.
Although this difficulty can be observed in the slightly decreasing accuracy  for values of $\fRealized>2.2$, which has been obtained by the  CNN-based approach proposed in the present paper, a satisfying accuracy was achieved even for high fractal dimensions,  as shown in Figure~\ref{result_fractal_dimension}a. Furthermore, the CNN approach works well independently of the aggregate size, see Figure~\ref{effect_aggregate_size}a. 

Probably the most comparable conventional method for determining the mixing ratio of an aggregate via its 2D STEM image, is based on determining the particle label of each pixel using a threshold value. More specifically, depending on the pixel intensity, the pixel is classified as \ce{TiO2}, \ce{WO3} or background, and then, using  the a priori known particle size distributions, a mixing ratio can be predicted.
However, since the representation of thick TiO$_2$ particles or of many 
 overlapping  TiO$_2$ particles  can have the same pixel intensity values as the representation of thin WO$_3$ particles, this threshold approach has a large source of errors~\cite{Gerken2023}.
The best appearing thresholds  using a ''brute force''  algorithm on a representative data were determined. This results in an MAE of 0.078 per aggregate when estimating the mixing ratio. Compared to the MAE of 0.053, see Section\ref{sec:result_mixing}, of the CNN approach described in the present paper, the error increases by 40\% for the  thresholding method described above.
 This is likely due to the increased values of pixel intensity  which are caused by overlapping particles (see Figure~\ref{effect of invert}), where these pixels with increased intensity values tend to be classified as \ce{WO3}.
 Therefore, conventional threshold methods become increasingly inaccurate with an increasing number of overlapping particles, contrary to the behavior of the CNN approach proposed in the present paper, see Figure~\ref{effect_aggregate_size}b.

Regarding the prediction of the remaining two model parameters $\cAModel$ and $\cBModel$, as far as we know, there is no comparable conventional method based on 2D image data.  Such  methods, if they do not consider depth information, would not be able to recognize if overlapping particles are touching or not, and thus, it is unlikely that they can accurately predict the values of $\cAModel$ and $\cBModel$.

\begin{figure}[ht]
    \centering
    \includegraphics[width=.5\textwidth]{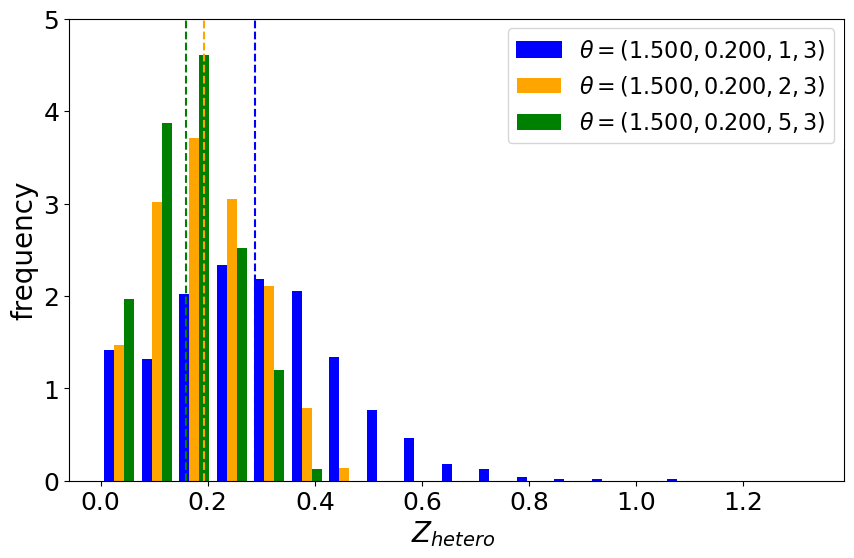}
    \caption{Distribution of the average heterogeneous coordination number
$\coordinationHetero(A)$ for three different values of
 $\cAModel$.  The dashed vertical lines show the  mean value of  $\coordinationHetero(A)$  for each of the three specifications of $\cAModel$, computed for 
    a total of 2400 simulated aggregates $A$. }
    \label{auswirkungen_clusterisze}
\end{figure}

Recall that the objective of the present paper is to generate digital shadows that are stochastically equivalent to the ground-truth aggregates used for model fitting. These digital shadows, which have known a 3D structure, can then be employed to predict the structural properties of the ground-truth aggregates at significantly reduced costs. 
Therefore, rather than just evaluating the accuracy of the predicted model parameters
$\widehat{\theta}=(\fModelpredicted,\rModelpredicted,\cAModelpredicted,\cBModelpredicted)$, the morphological similarities of the resulting digital shadows and their ground truth in terms of further structural descriptors, i.e.,  average clusters sizes and coordination numbers were also investigated.
As already mentioned in Section~\ref{sec.com.ori},
the MAE of $\widehat{\theta}$ is no appropriate tool to evaluate the similarity of  digital shadows and their ground-truth aggregates. 
For instance, an extreme mixing ratio leads to a situation where the precision of either $\cAModelpredicted$ or $\cBModelpredicted$ has only a negligible impact on the structure of the resulting aggregates due to the corresponding material occurring very rarely. Moreover, the structural similarity of the resulting digital shadows is more strongly affected by small errors and rounding of $\CNNcA(I(B))$ and $\CNNcB(I(B))$ when the values of $\cAModel$ and $\cBModel$ are small, as opposed to when they are large. In particular, errors in the prediction of ground truths for small values of  $\cAModel$ and $\cBModel$ result in higher relative errors. In such cases, large relative errors seem to have a greater impact on the structural discrepancies observed between aggregates generated for predicted and preset model parameters, see  Figure~\ref{auswirkungen_clusterisze}. This effect can be further exacerbated by the application of subsequent rounding operations.

Although the predictor 
$\widehat{\theta}=(\fModelpredicted,\rModelpredicted,\cAModelpredicted,\cBModelpredicted)$ proposed in this paper
shows only minor discrepancies across all descriptors listed in Table~\ref{tab::list_descriptors}, adapting the model and training process to address the issues mentioned above could enhance the similarity of  digital shadows and original aggregates even further. For example, expanding the possible values of $\cAModel$ and $\cBModel$ to the interval $[1,6]$, rather than just considering the discrete set $\{1, 2, \ldots, 6\}$, would result in  a diversity of aggregates, while also avoiding rounding errors that can arise in the prediction of $\cAModel$ and $\cBModel$.
More specifically, this could be achieved by modifying the aggregation model $\Psi_\numC$ introduced in Eq.~\eqref{def.psi.enn}, 
such that the sizes of the primary clusters are randomly distributed, instead of choosing a constant cluster size.  
This would achieve a more detailed coverage of possible aggregate structures, especially for small values of $\cAModel$ and $\cBModel$.
The training of CNNs could benefit from an adapted cost function that takes   the values of other model parameters into account and assigns weights to errors based on the importance of the  ground truth to be predicted.

\section{Conclusion}\label{sec.five}

     A method has been developed in order  to determine the parameters of a stochastic 3D model  for synthetic \ce{TiO2}-\ce{WO3} hetero-aggregates, based on their 2D STEM images. The method relies on convolutional neural networks that utilize distinct problem-specific architectures. 
     If such an appropriately calibrated stochastic 3D model is available, the neural network approach bypasses the need for using traditional microstructure analysis and modeling techniques, which are expensive in time and costs, such as tomographic STEM imaging as well as complex image processing and segmentation. 
     The networks were capable of predicting model parameters that describe fractal dimension, number-wise mixing ratio, and the sizes of primary clusters.
     The  aggregates drawn from the stochastic 3D model with predicted model parameters exhibited almost the same coordination numbers and average cluster sizes as those generated by the  model with the original (preset)  parameters.

    In the present paper, synthetic \ce{TiO2}-\ce{WO3} hetero-aggregates are used as  model system, because these two materials show a good material contrast in STEM images. However,  only spherical particles were used  for the generation of  synthetic 3D aggregates. It would be interesting to investigate the effectiveness of the proposed method if   particles for the hetero-aggregates are considered, which feature similar STEM intensities but differ significantly in their shape or size. Moreover, since experimentally measured aggregates feature more varied cluster sizes than  synthetically generated ones, it can be presumed that a larger variability in cluster sizes would require  more comprehensive data sets in order to make accurate predictions, but investigating this effect systematically is still important. Finally, in a forthcoming study, the presented method will be experimentally validated. More precisely, experimentally acquired 3D STEM image data of hetero-aggregates will be analyzed to investigate how well the  stochastic 3D model proposed on the present paper can describe real aggregates. 

\section{Acknowledgements}

This work was financially supported by the German Research Foundation (DFG)
through the research grants RO 2057/17-1, MA 3333/25-1 and SCHM 997/42-1.

\medskip

\bibliographystyle{abbrv}

\bibliography{refs}

\end{document}